\documentclass[12pt]{article}
\usepackage{setspace}
\usepackage{graphicx}
\usepackage{parselines} 
\usepackage{amsmath}
\usepackage{subfig}
\usepackage[left=25mm, right=25mm, top=25mm, bottom=20mm]{geometry}

\begin{document}  

\title{Towards the Self-constructive Brain: emergence of adaptive behavior}  

\author{Fernando Corbacho
\\~\\~\\Cognodata Consulting\\Paseo de la Castellana, 135. Madrid 28046 (Spain)
\\\&
\\Computer Science Department. Universidad Aut\'onoma de Madrid.
\\Carretera de Colmenar Viejo km. 15. Madrid 28049 (Spain)
\\~\\fernando.corbacho@cognodata.com}

\maketitle
\def\baselinestretch{1.7}

\tableofcontents

\newpage

\singlespacing

\begin{abstract}
\def\baselinestretch{1.7}

Adaptive behavior is mainly the result of adaptive brains. We go a step beyond and claim that the brain does not only adapt to its surrounding reality but rather, it builds itself up to constructs its own reality. That is, rather than just trying to passively understand its environment, the brain is the architect of its own reality in an active process where its internal models of the external world frame how its new interactions with the environment are assimilated. These internal models represent relevant predictive patterns of interaction all over the different brain structures: perceptual, sensorimotor, motor, etc. 
The emergence of adaptive behavior arises from this self-constructive nature of the brain, based on the following principles of organization: self-experimental,  self- growing, and self-repairing. Self-experimental, since to ensure survival, the self-constructive brain (SCB) is an active machine capable of performing experiments of its own interactions with the environment by mental simulation. Self-growing, since it dynamically and incrementally constructs internal structures in order to build a model of the world as it gathers statistics from its interactions with the environment. Self-repairing, since to survive the SCB must also be robust and  capable of finding ways to repair parts of previously working structures and hence re-construct a previous relevant pattern of activity. 
\end{abstract}

\def\baselinestretch{2}
\small

\begin{singlespacing}


\section{Introduction}
This paper claims that the emergence of adaptive behavior arises from the self-constructive nature of the brain. 
The brain builds itself up by reflecting on its particular interactions with the environment, 
that is, it constructs its own interpretation of reality through the construction of 
representations of relevant predictive patterns of interaction. 
Thus, in this paper we present the brain as the architect of its own reality. 
In this regard, we formalize the constructive architecture of the brain within the schema-based learning framework. 
We introduce the predictive (forward internal models) and its associated dual schemas as active processes capturing relevant patterns of interaction; and we suggest that the brain is composed of  myriads of these patterns.
These predictive internal models exist all over the brain and a variety of examples can be found in the literature regarding different brain areas/functionalities, for instance:
visual  (Berkes, Orban, Lengyel \& Fiser, 2011; Corbacho, 1997; Rao \& Ballard, 1999), 
sensorimotor (Corbacho \& Arbib, 1995; Mehta \& Schaal, 2002; Wolpert, Ghahramani \& Jordan, 1995), 
and motor (Corbacho \& Arbib, 1996; Corbacho, Nishikawa, Weerasuriya, Liaw \& Arbib, 2005b;  Desmurget \& Grafton, 2000; Flanagan \& Wing, 1997; Miall \& Wolpert, 1996; Shadmer, Smith \& Krakauer, 2010).
Yet,  these only represent the "tip of the iceberg', 
since they are all over all the different brain structures 
because they are the result of a central principle of organization
(Butz, 2008; Clark, 2013; Corbacho, 1997; Corbacho \& Arbib, 1997a,c; Downing, 2009; Grossberg, 2009; Haruno, Wolpert \& Kawato, 2001; Hawkins, 2004; Hoffmann, Berner, Butz, Herbort, Kiesel, Kunde \& Lenhard, 2007; Sigaud, Butz, Pezzulo \& Herbort, 2013; Wolpert \& Kawato, 1998). 
The existence of all these predictive internal models in the brain can be observed through many anticipatory activity patterns as reported for instance in (Bell, Han, Sugarawa \& Grant, 1997; 
Droulesz \& Berthoz, 1991; Duhamel, Colby \& Goldberg, 1992; Munoz, Pellison \& Guitton, 1991; Schutz \& Prinz, 2007; Suri \& Schultz, 2001; Szpunar, Watson \& McDermott, 2007). 
We claim that this anticipatory activity patterns are emergent of this type of constructive architecture, as it will be reported in this paper, showing anticipatory responses in sensorimotor and motor maps. 
This constructive architecture is also required to be able to construct and organize all these predictive patterns of interaction.

We propose that the brain is self-constructive (SCB) since it is self-experimental,          self- growing,         and self-repairing.
The brain is self-experimental since to ensure survival the self-constructive brain is an active machine capable of performing experiments of its own interactions with the environment 
as well as capable of mentally simulating the results of those interactions in order to be able to later decide the most optimal course of action. 
In this regard, the way for our brain to fully understand anything is to model and simulate it. 
To survive it must anticipate since anticipating an event allows to better prepare for it, since it allows the animal/agent to get ready to act immediately with the most successful course of action possible.
Anticipation  plays an important role in directing intelligent behavior under a fundamental hypothesis that 
the brain constructs reality as much as it embodies it. 
Hence, the anticipatory nature of the brain is a clear ingredient necessary for the SCB.
The brain is also self-growing, since it dynamically and incrementally constructs internal structures in order to build a model of the world as it gathers statistics while maintaining its energy consumption within bounds (Landauer principle). 
Finally, the brain is also self-repairing, since to survive, it must also be robust and capable of self-organization and self-repair, that is, the brain is capable of finding ways to repair parts of previously working structures and hence re-construct a previous successful pattern of interaction. 

These are evolutionary principles that have evolved in many different species, not just in mammals. That is, these principles can be clearly observed at different levels of abstraction and simple implementations can be  found even in lower vertebrates. 
So, for instance, anurans brain capacity for recovering a behavioral pattern of interaction after a critical lesion demonstrates their capacity for self-repair. Self-experimental and self-growing, on the other hand, can also be tested by their ability to learn certain problems such as learning to detour through the use of internal models. 
In this regard, this paper provides evidence in anurans and provides its computational counterpart in Rana Computatrix. 

The rest of the article is organized as follows. 
First we review the predictive brain, that is, how the brain builds anticipatory responses to expected patterns of interaction. 
We review the functionality of internal models as well as some experimental and computational data supporting their existence all over the brain. 
Then, we introduce the Schema-based learning (SBL) constructive architecture
as one possible computational  framework to formalize 
the constructive architecture of the SCB. 
It allows the formalization and construction of internal models by predictive schemas and their dual schemas, 
and through these computational models, is able to demonstrate how all the experimental data on learning prey-catching, later presented in this paper, can be explained following these previously mentioned SCB principles of organization.  

In the next section, we introduce prey-catching as it is one of the most critical behaviors for survival since feeding is clearly necessary to maintain vital constants.
We also study this behavior since it is evolutionary critical for all species. 
It is also an adaptive behavior since, in a changing dynamic environment, the conditions and  contexts for prey-catching might change in unprecedented ways.
First, we review experimental data on learning to detour in section 3.1, 
and then in section 3.2 provide experimental data on learning prey-catching after a lesion.
Both emergent adaptive behaviors will prove to be excellent examples of the SCB principles of organization.  
Afterwords, we introduce Rana Computatrix, the computational frog, as an experimental testbed in which we can test and verify these same principles.
In order to do so, we model the two behavioral/neuroethological cases previously presented in prey-catching. 
Anurans, due to their simpler structure compared to mammals,  afford the construction of complete overall visuo-motor structures and behaviors. 
Also, Rana computatrix, a set of evolving models of adaptive visuomotor coordination  constructed during years of research, affords the design of new adaptive behaviors over the already previously constructed models. 
Hence, a complete model can be laid down and hence overall behaviors can be analyzed in detail. 

The next three sections will serve to prove the self-experimental, self-growing and self-repairing principles of the SCB by explaining behavioral data in anurans under the schema-based learning constructive framework.
The self-constructive brain is self-experimental since to ensure survival the brain is an active machine capable of performing experiments of its own interactions with the environment 
as well as capable of mentally simulating the results of those interactions in order to be able to later decide the most optimal course of action. 
In order to understand its relation with the environment, the brain constructs internal models of reality that allow the brain to internally simulate its interactions with the environment. 
Hence, the brain has the capacity to design experiments to be performed by  interacting with the environment as well as  experiments inside the mind, for instance by mental practice/modeling, visualization and simulation. 
In this regard, cause-effect experimentation (section 5.1) triggers the construction of predictive schemas and its dual  schemas.  Hence, a very important component of the self-experimental brain consists on the ability to simulate the causal flow of the interactions with the environment. 
In this regard, predictive schemas allow for mental modeling. That is, they allow the system to anticipate the results of an action before it is taken. That is, predictive schemas produce anticipatory representations of events/effects that have no yet occurred in the external environment. 

The self-constructive brain is self-growing. In this regard we show an example of schema construction growing a new topological configuration to represent, and hence be able to reproduce, a successful relevant pattern of interaction. 
We show how the brain builds itself up using the SBL formalization of schema construction. 
We finish up by showing that the self-constructive brain is self-repairing.
Thar is, the brain has the capacity to repair itself and re-construct past functionality after certain lesions may take place. 
In order to so, we show how the constructive principles developed so far are able to explain learning after a lesion of the Hypoglossal nerve that initially prevents mouth opening during prey catching.

\section{The predictive brain: anticipatory construction of reality}
The brain is the architect of its own reality in an active process where its internal models of external reality frame how its new interactions with the environment are assimilated. We claim that these internal models represent relevant predictive patterns of interaction all over the different brain structures: perceptual, sensorimotor, motor, etc. 
In order to survive, the brain must anticipate since anticipating an event allows to better prepare for it, that is, it gets the animal/agent ready to act immediately with the most successful course of action possible (Sutton \& Barto 1981).
Hence, the anticipatory nature of the brain is a clear ingredient necessary for the SCB.

A key component of the predictive brain are internal models of the world, not only to understand but to actually construct its own reality. 
They allow the agent to anticipate the results of its own interactions with the environment and they play an important role in directing intelligent behavior under a fundamental hypothesis that 
the brain constructs reality as much as it embodies it. 
Many theories argue that the mind is for anticipation 
(Bar, 2007; 
Bubic, von Cramon, \& Schubotz, 2010;
Butz, 2008; Clark, 2013; 
Corbacho \& Arbib, 1997a; 
De Ridder, Verplaetse\& Vanneste, 2013;
Gilbert \& Wilson, 2007; Hawkins, 2004; Hoffmann, Berner, Butz, Herbort, Kiesel, Kunde \& Lenhard, 2007; Pezzulo, 2008; Pezzulo, Butz, Sigaud \& Baldassarra, 2009; Rosen, 1985), or more precisely, for building and working upon anticipatory representations (Corbacho, 1997; Miceli \& Castelfranchi, 2002; Hassabis \& Maguire, 2009). 
A real “mental” activity and representation begins to exist when the organism is able to endogenously (not as the output of current perceptual stimuli) produce an internal perceptual representation of the world -simulation of perception- (Castelfranchi, 2005).
Rather than passively “waiting” to be activated by sensations, it is proposed that the human brain is continuously busy generating predictions that approximate the relevant future (Bar, 2007). That is, the brain is proactive in that it regularly anticipates the future. 
Bar (2007) proposes that rudimentary information is extracted rapidly from the input to derive analogies linking that input with representations in memory. These predictions facilitate perception and cognition by pre-sensitizing relevant representations. 

Butz \& Hoffman (2002) explain how to use the perceptual expectation for implicitly monitoring the success of a rule-based reactive behavior, and as criteria for reinforcing or not a specific rule. But it can also entertain a mental representation of the current world just for working on it, modifying this representation for virtually exploring possible actions, events, simulations, and so on.
Intelligence is the capacity to solve a problem by working on an internal representation of the problem, by acting upon images with simulated actions, or on mental models by mental actions, transformations (reasoning), before performing the actions in the world.
The ability that characterizes a mind is that of building representations of the non-existent, of what is not currently (yet) true, perceivable (Castelfranchi, 2005).
A fully developed mind is able to build never-seen scenes, new possible combinations of world elements never perceived. It is a process of building and creation (by simulation) not just memory retrieval. 

The ability to construct a hypothetical situation in one's imagination prior to it actually occurring may afford greater accuracy in predicting its eventual outcome. The recollection of past experiences is also considered to be a re-constructive process with memories recreated from their component parts (Hassabis \& Maguire, 2009). Construction, therefore, plays a critical role in allowing us to plan for the future and remember the past. Conceptually, construction can be broken down into a number of constituent processes although little is known about their neural correlates (Hassabis \& Maguire, 2009). Moreover, it has been suggested that some of these processes may be shared by a number of other cognitive functions including spatial navigation and imagination. Recently, novel paradigms have been developed that allow for the isolation and characterization of these underlying processes and their associated neuroanatomy. 

As already mentioned, the existence of all these predictive internal models in the brain can be observed through many anticipatory activity patterns as reported for instance in (Bell et al., 1997; 
Droulesz \& Berthoz, 1991; Duhamel et al., 1992; Munoz et al., 1991; Schutz \& Prinz, 2007; Suri \& Schultz, 2001; Szpunar, Watson \& McDermott, 2007). 
There is, on the other hand,  an extensive literature on the use of mental models in cognitive architectures (e.g. Johnson-Laird, 1983).
Miall and Wolpert (1996) suggested that the brain internally simulates the behavior of the motor system in planning, control and learning. Such an internal “forward” model is a representation of the motor system that uses the current sate of the motor system and motor command to predict the next state. Miall and Wolpert (1996) outline the use of such internal models for solving several fundamental computational problems in motor control and review the evidence for the existence and use by the brain.
In the next section we provide a brief review on the literature of internal models in the brain.

\subsection{Internal models in the brain}
The importance of internal models in the brain has been acknowledged for many years
(e.g. Craik 1943; Gregory, 1967; Arbib 1972).
Internal models predict the evolution of the environment by imitating its causal flaw.
They play an important role in directing intelligent behavior under a fundamental hypothesis that 
the brain constructs reality as much as it embodies it. 
At the behavioral level we see that animals learn to anticipate predictable events. 
The term internal model is also popular in control theory
and denotes a set of equations that describes the temporal development
of a real world process (e.g. Kalman 1960; Garcia et al. 1989).

The concept of an internal model, a system which mimics the behaviour of a natural process, has emerged as an important theoretical concept in motor control (Jordan, 1983; Kawato, 1990, 1999; Kawato, Furukawa \& Suzuki, 1987;   Miall, 1993, 2003; Miall \& Wolpert, 1996; Wolpert \& Flanagan, 2001; Wolpert, Miall \& Kawato, 1998). 
Internal models can be classified into several conceptually distinct classes. 
One type of internal model is a causal representation of the motor apparatus, sometimes known as forward model (Jordan \& Rumelhart, 1992). Such a model would aim to mimic or represent the normal behavior of the motor system in response to outgoing motor commands. 
A forward model is a key ingredient in a system that uses motor outflow (also called efference copy) to anticipate and cancel the sensory effects of movement. The internal sensory signal needed to cancel reafference has been labeled corollary discharge. 

Potential uses of forward models include: canceling sensory reafference, distal supervised learning, internal feedback to overcome time delays, state estimation, and state prediction for model predictive control and mental practice/planning (Miall \& Wolpert, 1996). 
More specifically, predictive internal models may allow to
transform errors between the desired and actual sensory outcome for
a movement into the corresponding error in motor command,
to resolve ambiguous information,
to synthesize information from disparate sources,
to combine efferent and afferent information,
to perform mental practice (internal simulation) to learn to select
between possible actions,
to perform  state estimation in order to anticipate and cancel sensory
effects of ones own actions (not to distract attention/resources),
to reduce the credit assignment space,
to contribute to reasoning and planning by forming predictive chains, 
and finally to estimate the outcome of an action and
use it before sensory feedback is available, 
when delays make feedback control too slow for rapid movements 
(Miall \& Wolpert, 1996). 

One fundamental problem which the brain  faces in the context of motor control is that the goal and outcome of a movement are often defined in task-related coordinates (Jordan \& Rumelhart, 1992). 
A basic problem, therefore, exists in translating these task-related (visual or auditory) goals and errors into the appropriate intrinsic signals (motor commands and motor errors) which are required to update the controller. The forward model can be used to estimate the motor errors during performance by backpropagation of sensory errors through the model. 
In this paper we will show how this problem of distal supervised learning generalizes beyond motor control and applies to many schemas all over the SCB. 

More recent models have paid closer attention to biological details/plausibility.
These internal models exist all over the brain and a variety of examples can be found in the literature regarding different brain areas/functionalities, for instance:
visual  (Berkes et al., 2011; Corbacho, 1997; Rao \& Ballard, 1999), 
sensorimotor (Corbacho \& Arbib, 1995; Mehta \& Schaal, 2002; Wolpert et al., 1995), 
and motor (Corbacho \& Arbib, 1996; Corbacho, Nishikawa, Weerasuriya, Liaw \& Arbib, 2005b;  Desmurget \& Grafton, 2000; Flanagan \& Wing, 1997; Miall \& Wolpert, 1996; Shadmer, Smith \& Krakauer,  2010).
Yet, these only represent the "tip of the iceberg', 
since they are all over different brain structures 
because they are the result of a central principle of organization
(Butz, 2008; Corbacho, 1997; Corbacho \& Arbib, 1997a,c; Downing, 2009; Haruno, Wolpert \& Kawato, 2001; Hawkins, 2004; Hoffmann, Berner, Butz, Herbort, Kiesel, Kunde \& Lenhard, 2007; Wolpert \& Kawato, 1998). 
Anticipatory activity seems to reflect the processing of an internal model as already discussed in the previous section.
In the brain, anticipatory activity has been reported in the hippocampus,
anterior thalamus, frontal eye fields, superior colliculus, parietal cortex, 
striatum, and midbrain (see Corbacho, 1997; Downing, 2009 for reviews).

Another kind of internal models are known as inverse models (Atkeson, 1989), which invert the causal flow of the motor system. They generate, from inputs about its state and state transitions, an output representing the causal events that produced that state (Cruse \& Steinkuelher, 1993; Wada \& Kawato, 1993; Shadmehr \& Mussa-Ivaldi, 1994).
For example, an inverse dynamics model of the arm would estimate the motor command that caused a particular movement. The input might therefore be the current and the desired state of the arm; the output would be the motor command which would cause the arm to shift from the current state to the desired state. An inverse sensory output model would predict the changes in state that corresponded to a change in sensory inflow. 
In the kinematic domain the inverse kinematic model again inverts the forward kinematic model to produce a set of joint angles which achieve a particular hand position. However, as a forward model may have a many-to-one mapping, there is no guarantee that a unique inverse will exist. 

Historically, the notion of "internal models of the world" led
to the notion of schema (Arbib, 1972). 
Hence, we will use schema theory and particularly its extension schema-based learning (Corbacho \& Arbib, 1997c) as the framework to formalize the SCB.

\section{Schema-based Learning constructive architecture}
Schema theory and its extension with Schema-based learning (SBL from now on) might serve as a possible framework to formalize the SCB. As we have already mentioned, historically the notion of "internal models of the world" led to the notion of schema (Arbib, 1972). 
We first introduce and formalize the notion of schema to describe the modular functional and structural characteristics of the brain.
Then we introduce the notion of predictive schema to generalize and formalize the concept of internal forward models in the brain within the SBL architecture (Corbacho and Arbib 1997c; 
Corbacho, 1998).
The work presented here emphasizes the use and construction of 
new predictive internal models (predictive schemas) as well as 
the construction of their corresponding dual schemas when  specific conditions arise during the 
interaction of the agent with its environment. 
SBL also presents a more general 
approach including a wider variety of schemas, 
and a larger repertoire of processes to construct the different kinds of schemas
under various conditions. 

Corbacho and Arbib (1997a) introduced the notion of 
{\bf coherence}
to emphasize the importance of maximizing the  congruence between 
the results of an interaction (external or internal) and the expectations
(previously learned) for that interaction. 
The fundamental two principles of organization  in SBL are coherence and performance maximization.
Besides the main units of organization, the predictive schema and its dual associated schema, goal schemas are the other component in charge of dynamically setting a hierarchy of goals. 
SBL attempts to reduce incoherences and get closer to goal states simultaneously.
One of the main operations in SBL is the construction of all these internal models. 

We will briefly review the
notion of schema introduced in Corbacho (1997). 
We  described a schema  as a
unit of concurrent processing corresponding to a domain of interaction.
Lyons \& Arbib (1989) provided a formal semantics based on port
automata and Corbacho (1997) extended this definition to include 
schema activity variables and their dynamics. Other schema formalizations related to developmental learning are possible (e.g. Drescher, 1991). 
\begin{verbatim}

\end{verbatim}
\newpage
{\bf Definition:} A basic {\it schema} description is 
\begin{verbatim}
  basic-schema::= [Schema-Name:   <N>
  Input-Port-List:               (<Iplist>)
  Output-Port-List:              (<Oplist>)
  Variable-List:                 (<Varlist>)
  Behavior:                      (<Behavior>)]
\end{verbatim}	
where
$N$ is an identifying name for the schema $S^N$,

$<Iplist>$ and $<Oplist>$ are lists of $(Portname: Porttype)$ pairs
for input and output ports, respectively.

$<Varlist>$ is a list of $(Varname:Vartype)$ pairs for all internal
variable names, and 

$<Behavior>$ is a specification of computing behavior. 
\begin{verbatim}

\end{verbatim}

The notation $i^{x}_k(t)$ is used to represent the
pattern of activity in the $k$th
input port of schema $S^{x}$ at time $t$ and
  $o^{x}_k(t)$ for the analogous output port.
In the rest of the paper, in order to alleviate the use of notation,
when the schema has a single output port, then this output port takes the name of the
schema in lower case. That is, for instance we will use the notation $prey(t)$  to
name the output port activity pattern from the $S^{PREY}$ prey recognition visual schema. 

As we have already mentioned, Corbacho \& Arbib (1997c) also
introduced two special kinds of schemas, namely {\it predictive
schemas} and {\it goal schemas}. 
Goal schemas are a special kind
of schemas whose output port corresponds to a goal state (i.e. desired
pattern of activation in another schema). 
On the other hand, the role of the {\it predictive schema} is to anticipate the effect that the
particular pattern of activation of the {\it cause} schema has on the
current state of the {\it effect} schema. 

\subsection{Predictive and Dual schemas}
Corbacho (1997) presented predictive schemas as a generalization of
forward models (Jordan \& Rumelhart, 1992) since they not only apply to motor control but rather to perceptual, sensorimotor and abstract representational spaces as well. 
As already described, the role of the {\it predictive schema} is to anticipate the effect that the
particular pattern of activation of the {\it cause} schema has on the
current state of the {\it effect} schema. 
Also every predictive schema has
an associated {\it dual schema} that is responsible for selecting the optimal pattern of activity in the cause schema such that a goal pattern of activity in the effect schema can be achieved. 
\begin{verbatim}
\end{verbatim}
{\bf Definition:} A {\it predictive schema} $P^{x,y}$ associated with 
effect schema $S^x$ and cause schema $S^y$ is a schema with the
following special characteristics:

$<Iplist>$: The first input port is of the same type and connects to the
output port of the effect schema $S^x$, i.e. $i^{x,y}_1(t)=o^{x}(t)$.
The second input port is of the same type and connects to the
output port of the cause schema $S^y$, i.e. $i^{x,y}_2(t)=o^{y}(t)$.
The remaining input ports are optional and correspond to {\it context}
 information; i. e. $i^{x,y}_3(t)=o^{v}(t)$

$<Oplist>$: The output port contains the predictive response
$\hat{o}^{x,y}(t+1)$, representing the expectation for the state of the output port of the effect schema $S^x$ at time $t+1$, i.e., 
$o^{x}(t+1)$.  		

$<Varlist>$: Includes the parameters of the predictive schema, namely ${\bf W}^{x,y}_P (t)$.	

$<Behavior>$: The predictive schema behavioral specification includes a mapping $M^{x,y}_P$ parameterized by ${\bf W}^{x,y}_{P} (t)$, such that 
\begin{equation} 	
\hat{o}^{x,y}(t+1) = M^{x,y}_P (o^{x}(t), o^{y}(t), o^{v}(t); 
\newline 
{\bf W}^{x,y}_{P} (t))
\end{equation} 
 as well as a mapping $T_P$, 
\begin{equation}
{\bf W}^{x,y}_P (t+1)=T_P ({\bf W}^{x,y}_P (t), \\
\hat{o}^{x,y}(t+1), 
\\o^{x}(t+1) )
\end{equation}
$T_P$ allows the predictive schema to be tuned, i.e. its parameters 
change according to the prediction error $(o^{x}(t+1) - \hat{o}^{x,y}(t+1))$,
so that the predictive response
$\hat{o}^{x,y}(t+1)$
becomes increasingly closer to the observed response 
$o^{x}(t+1)$
as the number of interactions increases. 
In this regard, several error minimization methods
are valid. Also, depending on the architectural implementation of the schema, 
${\bf W}^{x,y}_P$ takes the corresponding form.
Corbacho et al. (2005) followed a particular neural network
implementation of both mappings $M^{x,y}_P (t)$ and $T_P (t)$ by a learning spatio-temporal mapping algorithm. Sanchez-Montanes and Corbacho (2004) presented an information theoretic metric to build this type of mappings.

Associated to the predictive schema $P^{x,y}$ there is always its corresponding dual schema $S^{y,x}$.
The predictive schema is the analogous to the forward internal models and the dual schema is the analogous to the inverse internal model (Jordan \& Rumelhart, 1992).
\begin{verbatim}
\end{verbatim}
{\bf Definition:} A {\it dual schema} $S^{y,x}$ associated with predictive schema $P^{y,x}$, 
produces the necessary optimal pattern of activity in cause schema $S^y$ 
in order to successively achieve the desired goal pattern of activity in the effect schema $S^x$.
It is a schema with the following special characteristics:

$<Iplist>$: The first input port is of the same type and connects to the
output port of the effect schema $S^x$, i.e. $i^{y,x}_1(t)=o^{x}(t)$.
The second input port is of the same type and connects to the
output port of the goal schema $G^{z,x}$, i.e. $i^{x,y}_2(t)=o^{*z,x}(t+1)$,
that is, a goal schema that is in charge of producing a goal pattern of activity for the effect schema $S^x$ 
(defined below).
The remaining input ports are optional and correspond to {\it context}
 information; i. e. $i^{y,x}_3(t)=o^{v}(t)$

$<Oplist>$: The output port is of the same type and connects to the input port of cause schema $S^y$ 

$<Varlist>$: Includes the parameters of the dual schema, namely ${\bf W}^{y,x}$.	

$<Behavior>$: The dual schema behavioral specification includes a mapping $M^{y,x}$ parameterized by ${\bf W}^{y,x}(t)$, such that
\begin{equation} 	
o^{y,x}(t)=M^{y,x}(o^{x}(t), o^{*z,x}(t+1), o^v(t); {\bf W}^{y,x}(t) )
\end{equation} 
as well as a mapping $T$, 
\begin{equation}
{\bf W}^{y,x}(t+1)=T({\bf W}^{y,x}(t), 
 o^{x}(t+1), 
 o^{*z,x}(t+1),
\frac{\partial \hat{o}^{x,y}}{\partial o^y})
\end{equation}
$T$ allows the dual schema to be tuned, i.e. its parameters 
change according to the performance error $(o^{*z,x}(t+1) - o^{x}(t+1))$,
so that the observed response $o^{x}(t + 1)$
becomes increasingly closer to the desired goal response $o^{*z,x}(t + 1)$ as the number of interactions increases. 
The problem of training the dual schema is a  {\it distal learning} problem (Jordan \& Rumelhart, 1992) since the parameters of the dual schema must be adapted based on the error on a distal space. That is, the dual schema must find parameters that recover the optimal patterns in the space of the cause schema, that is in the representation space of  $o^{y}(t)$ 
so as to reduce the difference in the distal error space $(o^{*z,x}(t+1) -  o^{x}(t+1))$. 
As Jordan and Rumelhart (1992)  described, and online learning algorithm based on stochastic gradient descend can be used. To perform adaptation, the change of weights must take into account  $\frac{\partial o^{x
}}{\partial o^y}$. Yet, the dependence of $o^{x}$ on $o^y$ is assumed to be unknown a priori. Yet, given a differentiable predictive forward model, it can be approximated by $\frac{\partial \hat{o}^{x,y}}{\partial o^y}$. That is, the distal error is propagated backward though the predictive schema (forward model) and down into the dual schema (inverse model) where the weights are actually changed accordingly\footnote{For the sake of clarity and simplification for this article, we assume that only one predictive and its corresponding dual schema are instantiated for each specific active motor schema. Hence,  avoiding issues of integration (linear on nonlinear). Thus, $\hat{o}^x(t) = \hat{o}^{x,y}(t)$, $o^y(t) = o^{y,x}(t)$ and $\frac{\partial \hat{o}^x}{\partial o^y} = \frac{\partial \hat{o}^{x,y}}{\partial o{y,x}} $.
In this regard, equation 19 will also be simplified to reflect this assumption.}. Kawato's (1990a) feedback-error-learning can also be used in this context. Later, Figure 5 will schematize this information flow once a Rana Computatrix model is introduced and described.

\subsection{Goal schemas and goal-oriented behavior}
Goal oriented behavior is one of the hallmarks of intelligent systems.
That is, the ability to set and achieve a wide range of goals
 (desirable states defining an objective signal).
Goal states must be stored so that they can be actively sought for in the future.
We must distinguish between implicit (hardwired) goals and learned goals and subgoals.
Drive reduction pertains to the first kind.
Where {\bf drives} can be viewed as states (Milner, 1977), which influence neurons either
 mechanically or chemically, and which have representations. Based on this hypothesis,
Arbib \& Lieblich (1977)
introduced a set $(d_i,...)$ of discrete drives to control the agent behavior. 
At each time $t$, each drive $d_i$ has a value $d_i(t)$. Drives can be appetitive or aversive.
Each appetitive drive spontaneously increases with time towards $d_{max}$, while
 aversive drives are reduced towards 0, both according to a factor $\alpha_d$. 
An additional increase occurs if an incentive $I(d,x,t)$ is present such as the sight or aroma of food
 (e.g. $S^{PREY}$ schema active) in the case of hunger.
Drive reduction $a(d,x,t)$ takes place in the presence of some
substrate -food reduces the hunger drive. If the agent is in the situation $x$ at time $t$, then the value of $d$ at time $t+1$ will become

\begin{equation} 
d(t+1)=d(t)+\alpha_d|d_{max}-d(t)|-a(d,x,t)|d(t)| \\
+I(d,x,t)|d_{max}-d(t)|
\end{equation} 

Internal drives (variables) must be kept within a restricted interval to assure
the survivability of the agent (intrinsic/ hardwired goals).
From this a hierarchy of subgoals has to be learned. That is, what states take the 
system to the primary goals (Guazzelli, Corbacho, Bota \& Arbib, 1998). 
Corbacho (1997) included $hunger$, $fear$, $thirst$, etc; 
in this paper we have included just $hunger$.
A goal corresponds to a state with a high drive reduction or in the way 
(anticipating drive reduction).
$x$ in the definition above is the state of the output port of a schema (or schemas)
at time $t$.
Corbacho (1997) described reinforcement type algorithms to store primary sensory goal states, that is, sensory states associated with high reward or anticipation of reward. 
These are specially needed in stochastic environments with delayed reinforcement (Sutton, 1988; 1990; Sutton \& Barto, 1998), more details in Corbacho (1997).

Hence, the ability to set and generate goals and subgoals is critical, such as the goal state of the jaw muscle  spindles indicating that the mouth must be open in order to get the prey. 
So the constructive brain architecture must be able to restore desired states so that they can be actively pursued. 
The desired state in a particular schema is triggered by the contextual state defined by the activity pattern in another schema. 
For secondary goals, goal states must be parametrized by contextual information.
That is, the goal pattern of activity must be produced by an adaptive mapping. 
So, for instance, prey-catching reduces the hunger drive,
which is signaled by the prey in the mouth,
so that when a prey is within the visual field a subgoal must be generated so that the jaw muscle spindles must get activated, indicating that the mouth is successefully open in order to allow the prey to get into the mouth.
During leaning, the reinforcement signals enhance the elegibility of the projections from the prey recognition schema to the jaw spindles goal pattern of activity representation (Corbacho, 1997). 
Another example in Rana Computatrix corresponds to the goal schema  that 
produces a goal pattern of representation in the motor heading map in the presence of prey, namely representing that in order to capture the prey it must be centered within its sensorimotor representation, that is the prey is "within grasp". 
Sections  4.3.1 and 4.3.2 will detail these goal schemas learned during learning to detour and learning to snap in Rana Computatrix respectively. 
\begin{verbatim}
\end{verbatim}
{\bf Definition:} A {\it goal schema} $G^{z,x}$ associated to (source) schema $S^z$ 
and (objective) schema $S^x$ (typically an effect schema in a predictive schema) 
is a schema with the
following special characteristics:

$<Iplist>$: The first input port is of the same type and connects to the 
output port of the source schema $S^z$, i.e. $i^{z,x}_1(t)=o^{z}(t)$.
The remaining input ports are optional and correspond to {\it contextual}
 information.

$<Oplist>$: The output port contains the {\it objective} response
${o}^{* z,x}(t+1)$, representing the desired pattern of activation in $o^{x}(t+1)$.  	

$<Varlist>$: Includes parameters of the goal schema, ${\bf {W}}^{z,x}_G$.	

$<Behavior>$: Implements a mapping $G$ from different inputs at different
times to the objective response, 
\begin{equation} 	
{o}^{* z,x}(t+1)=M^{z,x}_G( o^{z}(t)   ; {\bf W}^{z,x}_G (t) )
\end{equation} 
\noindent as well as a mapping $T_G$,
\begin{equation}
{\bf W}^{z,x}_G (t+1)=T_G({\bf {W}}^{z,x}_G (t+1),
{o}^{* z,x}(t+1), 
o^{x}(t+1))
\end{equation}
corresponding to the parameter tuning.
A specific spatio-temporal mapping learning algorithm to learn this mapping is
presented in (Corbacho et al., 2005b) and a generalized information-theoretic measure is presented in (Sanchez-Montanes \& Corbacho, 2004).

\section {Adaptive prey-catching behavior in Anurans and  Rana Computatrix}
Prey-catching is one of the most critical behaviors for survival since feeding is clearly necessary to maintain the animal's vital constants.
We study this behavior since it is evolutionary critical for all species. 
It is also an adaptive behavior since, in a changing dynamic environment, the conditions and  contexts for prey-catching might change in unprecedented ways. In this regard we describe two adaptive patterns of interaction to catch prey, namely learning to detour around a barrier to approach the prey and learning to snap the prey after the lesion of the Hypoglossal nerve has occurred.
Innate motor synergies allow the froglet to snap quite precisely to moving objects (Weerasuriya and Licata 1996).
Yet more precise and discriminative behaviors require learning from the environment. 
That is the case when anurans learn not to snap to large objects by interacting with preys of different sizes (Weerasuriya and Licata 1996). 
On the other hand, the animal might also suffer different types of lessions during its lifetime and recovering its prey-catching capabilities under different structural changes is obviously critical for its survival. 

This section briefly introduces experimental data on anurans and then, the later subsection, describes models for these behaviors within the computational framework of Rana Computatrix.
As already expressed, Rana Computatrix consists of a set of evolving models of anuran adaptive
visuomotor coordination (Arbib, 1987). Corbacho and Arbib (1995) and Corbacho, Nishikawa, Weerasuriya,  Liaw and Arbib (2005a,b) emphasized adaptive prey-catching behavior in Rana computatrix.

\subsection{Learning to Detour: Experimental Data}
Frogs have an innate ability to detour around narrow barriers (Collet, 1982). That is, when frogs are presented with a barrier 10-cm wide made of a pailing fence, the frog is able to detour to catch a worm on the other side of the fence (Figure 1A).
A plausible explanation is due to the motor heading map (mhm) integration hypothesis 
by which the prey produces an attractant field which is is integrated with a repellent field produced by the obstacles (Corbacho \& Arbib, 1995). 
For the case of the narrow 10-cm. wide pailing fence the prey attractant field extends beyond the narrow barrier repellent field, as it will be later explained. 
On the other hand, for wider barriers (i.e. 20-cm wide) frogs are uncapable of detouring around the fence and they actually go straight to the position just in front of the prey and try to catch the prey by pushing the fence in repeated occasions (Figure 1B). 
\begin{verbatim}



\end{verbatim}

\begin{center}
----------- FIGURE 1 ABOUT HERE -----------
\end{center}
Yet after several trials interacting with the wide barrier, the frogs learn to detour around the barrier, so that when placed at the start position, after training,  they immediately produce a series of sidestep and forward moves that completely avoid the barrier as shown in Figure 1C
(Corbacho \& Arbib, 1995).

\subsection{Learning to Snap after Lesion: Experimental Data}
Initially,  following bilateral transections of the hypoglossal nerve, 
anurans
lunge toward mealworms with no accompanying tongue or jaw movement as displayed in Figure 2 (Weerasuriya 1989, 1991; Anderson and Nishikawa 1993).
This is so, since, before the lesion, the mouth opens thanks to the delay produced by the hypoglossal nerve (HG) signal to the levator muscles (LM) with respect to the depressor muscles (DM). 
Yet, following the lesion, they eventually 
learn to catch their prey by 
learning to open their jaw again and lunging their body farther (Innocenti and Nishikawa 1994;  Gleason and Nishikawa 1996;
Weerasuriya and Mills 1996). 
Innocenti and Nishikawa's (1994) as well as Gleason and Nishikawa (1996) studies on motor learning following hypoglossal transection in toads show that 
the individual learning process is quite idiosyncratic. 
That is, each toad seems to develop its own solution to the 
problem. The adaptation of the motor output may depend highly on the particular features of the prey. 
Thus, the distinction between jaw prehension and tongue prehension is only a qualitative analysis. 
Anurans use a visual 
analysis of prey features to modulate many aspects of motor output. Kinematic variables that displayed a 
learning effect showed a significant change during the motor learning period compared to the initial values 
(Innocenti and Nishikawa 1994).
Immediately following bilateral hypoglossal nerve transection, the percent of successful mouth opening, 
prey contact, and prey capture made by toads decreased. Yet all this improved by interacting with prey 
(Innocenti and Nishikawa 1994;  Gleason and Nishikawa 1996; Weerasuriya and Mills 1996). 
Before surgery, capture success was nearly 100\%. 
After surgery it fell to 0\%, 
and then returned to 100\% within 5-6 weeks. 
Innocenti and Nishikawa (1994) 
distinguish several learning stages as a result of different dependencies (e.g., to catch prey the animal must first 
learn to open the mouth) and number of "variables" involved. That is, reaching for prey involves more motor 
synergies than opening the mouth, there are also several possible ways of achieving it and more learning time is 
required to coordinate and adjust the overall synergy. Next let us summarize
the first stage, learning to open the mouth which is central for this article and appears to be less idiosyncratic.
\begin{verbatim}



\end{verbatim}

\begin{center}
----------- FIGURE 2 ABOUT HERE -----------
\end{center}
Innocenti and Nishikawa (1994)
observed that the first successful mouth opening for toad \#1 occurred on the 5th day after one practice session 
on the 16th trial following surgery. Toads \#2 and \#3 successfully opened their mouths on the day surgery was 
performed and on the 3rd day on the first practice session on trial 2 and 1, respectively.
Innocenti and Nishikawa (1994) and Gleason and Nishikawa (1996)
 also observe a general overshooting of mouth opening in the first 
learning trial after the lesion.
 In conclusion the first stage of learning to open the jaw appears to be common to all toads and, though there
are temporal differences during the learning process, some aspects are common 
(e.g. first overshooting of mouth opening). 
 After the lesion, the toad receives no direct external reinforcement from the environment. 
So we have claimed that an expectation-based strategy is one of the few viable ways the 
problem can be resolved (Corbacho, Nishikawa, Weerasuriya, Liaw \& Arbib, 2005a,b).  

\subsection{Constructive adaptive behaviors in Rana Computatrix}
As already mentioned, Rana Computatrix consists of a set of evolving models of anuran adaptive
visuomotor coordination (Arbib 1987). 
Corbacho and Arbib (1995) focused on the structures involved in the learning to detour behavior.
Corbacho (1997) provided a wider repertoire of behaviors in Rana Computatrix
and Corbacho et al. (2005) emphasize the structures for learning to snap prey-catching motor behavior after the Hypoglossal nerve lesion.
In this paper we generalize both cases within the SBL constructive architecture which is 
able to solve both problems with the exact same constructive machinery. 

\subsubsection{Learning to Detour in Rana Computatrix}
Corbacho and Arbib (1995) provided a basic schema-based model of learning to detour in Rana computatrix.
The initial  seed schema architecture allows for coarse interaction with the environment (Figure 3). 
During the detour behavior, the involved perceptual schemas are  $S^{PREY}$, $S^{SOR}$, $S^{TACTILE}$, the sensorimotor schemas are $S^{MHM}$, $S^{BUMP-AVOID}$,  and
the involved motor schemas are $S^{FORWARD}$, $S^{BACK-UP}$, $S^{SIDE}$, $S^{ORIENT}$ and  $S^{SNAP}$, which get activated at different times during the interaction of the agent with the environment (Figure 3). 
In this paper, we provide with only the strictly necessary details and refer the reader to the previous works for more detailed descriptions. 
The prey recognition schema $S^{PREY}$ gets activated when there is a prey within the visual field of the animat and sends an excitatory attractant field to the motor heading map represented in schema $S^{MHM}$. 
On the other hand, the stationary object recognition schema $S^{SOR}$, gets activated by the view of a stationary fence and sends an inhibitory repellent field to the motor heading map. Hence, the motor heading map schema $S^{MHM}$ integrates both fields to determine a course of action. 
\begin{verbatim}



\end{verbatim}

\begin{center}
----------- FIGURE 3 ABOUT HERE -----------
\end{center}
For the case of the animal avoiding the narrow pailing fence, as it has been already explained, a  plausible explanation is due to the motor heading map (MHM) integration hypothesis.
That is, when the prey attractant field goes beyond the narrow barrier repellent field (Figure 4A). 
Yet for wider barriers,  frogs are uncapable of avoiding the barrier and go forward against it. 
Again, a plausible explanation might be due to the fact that the prey attractant field can not extend beyond the wider repellent field projected by the 20-cm wide barrier as shown in Figure 4B. 
\begin{verbatim}



\end{verbatim}

\begin{center}
----------- FIGURE 4 ABOUT HERE -----------
\end{center}
Corbacho \& Arbib (1995) informally introduced the notion of relational predictive schemas and simply sketched out the construction of the detour schema under that context. 
In this regard, many predictive schemas imply a sensorimotor map and a motor action.
That is, they record the effect of the motor action over the dynamic representation in a specific sensorimotor map. 
The associations between the sensorimotor map schemas and the motor schemas $(S^x, S^y)$ are learned through cause-effect relations later explained in section 5.1 in this paper.
So, for instance, the activity pattern in the motor heading map $mhm(t)$ can be affected by different motor schemas and a different predictive schema is learned for each of this cause-effect relations. 

Before learning has occurred, the frog goes straight to the center of the 20-cm wide barrier, hence bumping into it. This tactile contact against the fences activates the $S^{BUMP-AVOID}$ schema which 
in turn activates the $S^{SIDE}$ sidestep schema to avoid contact with the pailing fence immediately in front of the agent. 
Eventually,  during the interaction with the wide barrier a new cause-effect relation is discovered between $(S^{MHM}, S^{SIDE})$ when the frog, after several sidesteps triggered by the  $S^{BUMP-AVOID}$ schema, happens to reach the end of the wide barrier and hence, 
the prey suddenly and unexpectedly appears openly within its visual field. 
That is, the unexpected pattern in the motor heading map, due to the view of the prey in the open field, triggers the new cause-effect relation $( S^{MHM}, S^{SIDE})$ which in turn will cause the construction of the following predictive and dual schemas:
\begin{equation}
\hat{mhm}(t+1) = M_P^{MHM, SIDE} (mhm(t),
side(t), sor(t))
\end{equation}
\begin{equation}
o^{SIDE, MHM}(t) = M^{SIDE, MHM}(mhm(t),
 mhm^*(t+1), sor(t))
\end{equation}

These two new schemas explain how the frog can learn to detour around the barrier after training with the 20-cm wide barrier has taken place. 
That is, after learning, when the frog is placed in front of the 20-cm wide barrier (the context) in the presence of prey,
 the sidestep schema $S^{SIDE}$ is activated since it receives input from the dual schema $S^{SIDE, MHM}$  (eq. 9) and hence the frog detours around the barrier. 
As already mentioned in section 3.2,  the goal schema $G^{PREY,MHM}$  
produces the goal pattern of representation in the motor heading map $mhm^*(t)$ in the presence of prey, 
namely, it represents that in order to capture the prey, it must be centered within its sensorimotor representation, 
that is the prey is "within grasp" ($prey(t) \rightarrow mhm^*(t)$). 
\begin{verbatim}



\end{verbatim}

\begin{center}
----------- FIGURE 5 ABOUT HERE -----------
\end{center}
In turn, when the dual schema is activated, its associated predictive schema is also activated and hence a predictive anticipatory pattern arises in the motor heading map, 
namely $\hat{mhm}(t+1)$ (see Figure 6). 
\begin{verbatim}



\end{verbatim}

\begin{center}
----------- FIGURE 6 ABOUT HERE -----------
\end{center}
To summarize the overall information flow depicted in Figure 5, it all starts with the activation of the schema $S^{PREY}$ due to the presence of a prey within the visual field of the frog; $prey(t)$ in turn activates the goal schema $G^{PREY,MHM}$, that produces the goal pattern of activity $mhm^*(t+1)$ for the motor heading map $S^{MHM}$.
This goal pattern $mhm^*(t+1)$ in conjunction with activation of the context by the activation of the schema $S^{SOR}$, in turn activates the dual schema $S^{SIDE, MHM}$ in charge of producing an input modulatory activity pattern $o^{SIDE, MHM}(t)$ for the motor schema $S^{SIDE}$ to try to achieve the goal activity pattern $mhm^*(t+1)$.
In parallel, the output activity pattern of the $S^{SIDE}$ schema also causes the activation of the predictive schema
$P^{MHM, SIDE}$  which, then, produces the anticipatory pattern of activity $\hat{mhm}(t+1)$ (Figure 6). 

In terms of the adaptation of the specific schemas, in order for the system to reach coherence with its environment (Corbacho \& Arbib, 1996), 
there are two types of errors (Jordan \& Rumelhart, 1992), the prediction error $(mhm(t+1) - \hat{mhm}(t+1))$, and the performance error  $(mhm^*(t+1) - mhm(t+1))$; responsible for tuning the predictive and the dual schemas respectively within a distal supervised learning paradigm. 
That is, the system faces a distal learning problem (Jordan \& Rumelhart, 1992)  since,
as already expressed, the parameters of the dual schema $S^{SIDE,MHM}$ are adapted based on the error on the distal motor heading sensorimotor space, namely based on the distal error $(mhm^*(t+1) - mhm(t+1))$. 

Following sections of this paper  will provide the general framework that allows the construction of these two schemas. That is, section 5.1 will provide the cause-effect constructive dynamics to be able to construct the association $(S^{MHM}, S^{SIDE})$ and sections 6.1 and 6.2 will explain the general constructive architecture that allows for the new schemas $P^{MHM,SIDE}$ and $S^{SIDE, MHM}$ to be constructed.

\subsubsection{Learning to Snap after the HG lesion in Rana Computrix}
Analogously, to learning to detour, the innate seed schema architecture allows for coarse interaction with the environment. 
Corbacho, Nishikawa, Weerasuriya, Liaw and Arbib (2005) provided a series of two papers, in paper I the schema-based  architecture was defined, and in paper II  schema-based learning after the HG lesion was described. 
In  this paper we only provide the strictly necessary abridged summary and refer the reader to the other papers for all the details on Rana Computatrix.
Figure 7 shows the initial seed schema-based architecture for learning to snap in Rana Computatrix.
It consists of perceptual schemas $S^{PREY}$, $S^{JAW\_REC}$, $S^{HG\_REC}$, $S^{TONGUE\_REC}$, $S^{DISTX}$, $S^{DISTY}$; 
sensorimotor schemas $S^{INT1}$, $S^{LM}$, $S^{DM}$, $S^{HO}$, $S^{GG}$, $S^{LU}$, $S^{HE}$; 
and motor schemas $S^{LEVATE}$, $S^{DEPRESS}$, $S^{PROTRACT}$, $S^{RETRACT}$, $S^{LUNGE}$, $S^{HEAD\_DOWN}$.
For the purpose of this paper only a subset of the schemas must be taken into account.
Namely, $S^{HG\_REC}$ is the proprioceptive schema reflecting the Hypoglossal nerve input signal.  
$S^{LM}$ and $S^{DM}$ represent premotor schemas for levator and depressor motor schemas,
$S^{LEVATE}$ and $S^{DEPRESS}$. 
During normal feeding behavior, due to the input signal from HG, the levator peak of activity is delayed with respect to the depressor peak of activity to allow for jaw opening as displayed in figure 2 bottom. 
As already specified, along this paper we will use the simplified notation $dm(t)$, $lm(t)$ and
$jaw\_rec(t)$ to mean the output port activity pattern from the $S^{DM}$,
$S^{LM}$ and $S^{JAW\_REC}$ schemas respectively.
\begin{verbatim}



\end{verbatim}

\begin{center}
----------- FIGURE 7 ABOUT HERE -----------
\end{center}
The temporal patterns of activity for every schema instance 
are displayed in Figure 8 when the animal is
catching prey before (Figure 8A) and after the HG lesion (Figure 8B).
They display catching a small prey with the tongue (coordinated control motor) pattern (TP) (left column in both figures) and 
catching a large prey with the jaw pattern (JP) (right column).
The tongue pattern produces a farther activation of the tongue whereas the jaw pattern catches larger prey by wider opening of the mouth and further lunging of the body (Corbacho, Nishikawa, Weerasuriya, Liaw \& Arbib,  2005a)
The reader should  pay special attention to the $S^{HG\_REC}$,
$S^{JAW\_REC}$ and $S^{LM}$ schemas activity patterns for the purposes of this paper 
(represented in rows 3, 4 and 12 respectively).
In this regard, the delay between the depressor peak of activity $dm(t)$ and the
levator peak of activity $lm(t)$ is related to body size. 
Thus, the control circuit
needs to be adaptive since the animal changes its size through its
lifetime. 
This points to the existence of internal adaptive models. 
So the system includes {\it internal models} which take efference
copies (von Holst \& Mittelstaedt, 1950) of the neural commands (e.g. premotor $S^{LM}$) 
and provide the expected effect of these commands.
The system also includes a variety of sensory feedback,
for flexible feedback control,
which are hereby shown to be useful for processes of adaptation as well (e.g. the jaw muscle spindles $S^{JAW\_REC}$). 
The expected feedback and the sensory feedback must be 
temporally matched to reach coherence. 
For instance, the signal for the open jaw is delayed, that
is, the muscle spindle feedback occurs 30 to 50 msec after {\it depressor} premotor
onset - mainly the time required for muscle contraction. 
Hence, a learned transformation (through a internal model) is involved to reach coherence between both signals since the efference copy 
will be different in both its structure and onset timing from the sensory feedback signal. 
\begin{verbatim}



\end{verbatim}

\begin{center}
----------- FIGURE 8 ABOUT HERE -----------
\end{center}
Another crucial feature is the use of incoherence signals to drive the learning process.
For example, we shall see how the system can detect incoherences by subtracting the signal carrying the 
efference copy about opening the mouth and the feedback signal from the jaw.
In this regard, in section 7.1 we shall show how internal models can be
learned and how the system can recover after a lesion by structural
learning thanks to the incoherences detected by the internal models.

During the lifetime of the animal (before the HG lesion)  prey-catching reduces the hunger drive,
which is signaled by the prey in the mouth,
so that when a prey is within the visual field a subgoal is  generated so that the jaw muscle spindles must eventually get activated, indicating that the mouth is successfully open in order to allow the prey to get into the mouth.
During goal learning, the reinforcement signals enhance the elegibility of the projections from the prey recognition schema to the jaw spindles goal pattern of activity representation (Corbacho, 1997). 
As already mentioned in section 3.2, the goal schema 
$G^{PREY,JAW\_REC}$ gives rise to the desired state of the jaw (spindles) when a 
particular prey is present, so that  ${o}^{PREY,JAW\_REC}(t) = {jaw\_rec}^*(t+1)$
represents the desired output activity pattern for $jaw\_rec(t+1)$.
Summarizing, this goal schema represents the pattern of interaction that the mouth must be open (subgoal) in order to be able to capture the prey (final goal associated to the hunger drive).

Also,  during the interaction with the prey, a new cause-effect relation is discovered between $(S^{LM}, S^{JAW\_REC})$ as it will be later explained in section 5.1.
In this regard, during normal snapping behavior a new pair of schemas is constructed (possibly among others), as it will be described in sections 6.2 and 7.1.
Namely, 
\begin{equation}
\hat{jaw\_rec}(t+1) = M_P^{JAW\_REC, LM} (jaw\_rec(t),
 lm(t))
\end{equation}
\begin{equation}
o^{LM, JAW\_REC}(t) = M^{LM, JAW\_REC}(jaw\_rec(t),
 jaw\_rec^*(t+1))
\end{equation}

These two schemas explain how the frog can learn to snap after the HG lesion has taken place.  
After the HG lesion  has occurred, the frog does not open the mouth and hence an incoherence occurs since the predictive schema $P^{JAW\_REC, LM}$ activates an expected pattern of activation in the $S^{JAW\_REC}$ schema, namely $\hat{jaw\_rec}(t+1)$. 
Also $jaw\_rec(t) \neq 0$ becomes a subgoal state since it is in the way to
reduce the hunger drive (the mouth must be open to capture the prey).
In this regard, the goal schema $G^{PREY,JAW\_REC}$ outputs
the desired state of $S^{JAW\_REC}$ when $S^{PREY}$ is active based on the parameters of the current prey (since $prey(t)$ is the incentive for hunger).
Hence $prey(t)$ activates and parametrizes the subgoal pattern of activity $jaw\_rec(t+1)^*$ which in turn serves as input for other schemas, namely as input for the dual schema $S^{LM, JAW\_REC}$.
\begin{verbatim}



\end{verbatim}

\begin{center}
----------- FIGURE 9 ABOUT HERE -----------
\end{center}
Figure 9 displays the information flow for the "chain" of schemas 
 $prey(t) \rightarrow {jaw\_rec}^*(t+1) \rightarrow {lm}(t)$
 triggered by the goal schema $G^{PREY,JAW\_REC}$ and 
the dual schema $S^{LM, JAW\_REC}$ respectively.
This figure has the same topological configuration and information flow as that of figure 5 for the case of learning to detour. 
Hence, proving the generality of the SBL constructive architecture for different behavioral cases. Thus, after learning, when a prey is placed within snapping distance of the frog,
the premotor LM schema $S^{LM}$ is activated since it receives modulatory input $o^{LM, JAW\_REC}(t)$ from the dual schema $S^{LM, JAW\_REC}$  (equation 11), 
which in turn was activated by the goal pattern of activity $jaw\_rec^*(t+1)$ (Figure 10). 
This new schema gives rise to the corrective (modulatory) pattern of activity in $lm(t)$
displayed in the third row in Figure 11.
and hence the frog is able to snap at the prey.
\begin{verbatim}



\end{verbatim}

\begin{center}
----------- FIGURE 10 ABOUT HERE -----------
\end{center}
In turn, when the dual schema is activated, its associated predictive schema is also activated and hence, a predictive anticipatory pattern arises in $\hat{jaw\_rec(t)}$.
Again, as for the learning to detour case,  it is a distal learning problem, since the parameters of the dual schema must be adapted based on the error produced on a distal sensorimotor space, namely $(jaw\_rec^*(t+1) - jaw\_rec(t+1))$. 
\begin{verbatim}



\end{verbatim}

\begin{center}
----------- FIGURE 11 ABOUT HERE -----------
\end{center}

\section{The Self-constructive brain is self-experimental}
The brain is self-experimental since to ensure survival the self-constructive brain is an active machine capable of performing experiments of its own interactions with the environment 
as well as capable of mentally simulating the results of those interactions in order to be able to later decide the most optimal course of action. 
In this regard, the way for our brain to fully understand anything is to model and simulate it. 
As Pezzulo (2008) has expressed, cognition is for doing, for simulating. 
In order to understand its relation with the environment the brain constructs internal models of reality that allow the brain to internally simulate its interactions with the environment. 
Hence, the brain has the capacity to design experiments to be performed by  interacting with the environment as well as  experiments inside the mind, for instance by mental practice, visualization and simulation. 

There is a growing body of experimental data that supports the idea of the self-experimental brain. 
In this regard,  experimental evidence indicates that animals can use mental simulation to make decisions about the action to take during goal-directed navigation (Chersi, Donnarumma, \& Pezzulo, 2013). Its most salient characteristic is that choices about actions are made by  simulating movements and their sensory effects using the same brain areas that are active during overt execution. 
Chersi et al. (2013) link these results with a general framework that sees the brain as a predictive device that can “detach” itself from the here-and-now of current perception using mechanisms such as episodic memories, motor and visual imagery. 
In this regard, the concept of action simulation is gaining momentum in cognitive science, neuroscience, and robotics, and in particular within the study of grounded, embodied and motor cognition (Declerck, 2013;  Hesslow, 2012; Raos, Evangeliou \& Saraki, 2007; Jeannerod, 2001; Mohan, Sandini \& Morasso, 2014; 
Pezzulo, Candini, Dindo, \& Barca, 2013). 

The ability to construct a hypothetical situation in one's imagination prior to it actually occurring may afford greater accuracy in predicting its eventual outcome. The recollection of past experiences is also considered to be a re-constructive process with memories recreated from their component parts (Hassabis \& Maguire, 2009). Construction, therefore, plays a critical role in allowing us to plan for the future and remember the past. Conceptually, construction can be broken down into a number of constituent processes although little is known about their neural correlates. Moreover, it has been suggested that some of these processes may be shared by a number of other cognitive functions including spatial navigation and imagination. Recently, novel paradigms have been developed that allow for the isolation and characterization of these underlying processes and their associated neuroanatomy. Hassabis \& Maguire (2009) selectively review this fast-growing literature and consider some implications for remembering the past and predicting the future.
In this regard, mental practice is the cognitive rehearsal of a physical skill in the absence of overt physical movement (Jordan, 1983). 
The questions arises whether mental practice enhances performance (Driskell, Copper \& Moran, 1994; Gentili, Han, Schweighofer \& Papaxanthis, 2010; Miall \& Wolpert, 1996).  
Mental practice promotes motor anticipation as for example there is evidence from skilled music performance.

On the other hand, mental imagery can be described as the maintenance and manipulation of perception and actions of a covert sort, i. e., it arises not as a consequence of environmental interaction but is created internally by the brain (Di Nuovo, Marocco, Di Nuovo \& Cangelosi; 2013; Kossyln, Gani \& Thompson, 2001; Lallee \& Dominey, 2013; Svensson, Thill \& Ziemke, 2013).
According to the simulation hypothesis, mental imagery can be explained in terms of predictive chains of simulated perceptions and actions, i.e., perceptions and actions are reactivated internally by our nervous system to be used in mental imagery and other cognitive phenomena (Chersi, Donnarumman \& Pezzulo, 2013; Hesslow, 2002, 2012; Svensson, Thill \& Ziemke, 2013). In this regard, Svensson, Thill \& Ziemke (2013) go a step further and  hypothesize that dreams informing the construction of simulations lead to faster development of good simulations during waking behavior.

Certain type of generative models such as the wake-sleep type algorithms (Hinton, Dayan, Frey \& Neal, 1995) are composed of forward connections as well as backward connections which can give rise to mental visualizations. 
That is, during the sleep generative phase, phantasy patterns corresponding to visualizations can be generated thanks to the information flow generated by the backward projection. 
In the same way, we show that predictive schemas are capable of producing anticipatory activity patterns as the result of mental modeling. 
In this regard, we have shown in figure 6 anticipatory (phantasy) activity pattern in the motor heading map ($\hat{mhm}(t+1)$) during learning to detour.
Also in figure 11, we have displayed  internally simulated activity patterns in pre-motor neurons 
($\hat{lm}(t+1)$) 
during learning after the hypoglossal nerve lesion. 
The prediction of states of the body has also been shown to be a useful capability in resilient robots (Bongard, Zykov, \& Lipson, 2006) and it has been suggested that this could be generalized to internal models of the environment (Adami, 2006). 

We have already introduced internal models in section 2.1 and their SBL implementation, predictive and dual schemas in section 3.1. Predictive schemas allow for mental modeling. 
That is, they allow the system to anticipate the results of an action before it is taken. That is, predictive schemas produce anticipatory representations of events/effects that have no yet occurred in the outside environment. 
That is, the self-experimental brain produces anticipatory experimental patterns of activity.
Yet the question remains as to how the predictive schemas are constructed. 
In this regard, cause-effect experimentation triggers the construction of predictive schemas and its corresponding dual schemas.  
Hence, a very important component of the self-experimental brain consists on the ability to simulate the causal flow of the interactions with the environment and, thus, learn cause-effect relations from the potentially combinatorially large set of hypothesis. 
In learning new causal relations by experimentation, the potentially combinatorial explosion is constrained by interacting with the environment. In an analogous manner as constraint relaxation dynamics reduce an exponential space into a more tractable space of possible configurations. 
Next, we will introduce the mechanisms to learn the effective 
cause-effect relations within a complex system (e.g. brain, autonomous agent).

\subsection{Constructing Cause-Effect relations}
In this section we propose learning mechanisms that will be able to
construct the cause-effect relations from the combinatorially
large relational space.  
As already indicated, these cause-effect relations will in turn allow for mental simulation and experimentation.
To decide on a cause-effect relation between
two schemas $S^x$ and $S^y$ we will use some measure
of {\it delayed similarity} of the activity patterns out of their
output ports during a time interval i.e. $o^x[t_a,t_b]$ and 
$o^y[t_a+\tau,t_b+\tau]$ respectively.
In general we will assume that any schema connects to any other
schema through several
connections, we assume that these different connections will have different
"delay properties", thus giving rise to different responses in
different connections. 
We now define the instantaneous degree of cause-effect relation between two patterns
$o^x(t)$ and $o^y(t)$, at a point in time $t$ by the
following measure

\begin{equation}
c^{x,y}_{\tau} (t) = \Theta[o^x(t)] \cdot \Theta[o^y(t)]-\alpha d^{x,y}_{\tau}(t) 
\end{equation}
\noindent where  
\begin{equation}
d^{x,y}_{\tau}(t) = (o^y(t-\tau) - o^x(t))^2
\end{equation}

  \[
    \Theta[x]=\left\{
                \begin{array}{ll}
                  1 & \mbox{if $x \neq 0$}\\
                  0 & \mbox{otherwise}\\
                \end{array}
              \right.
  \]

Two signals may not be very similar in their fine temporal structure
yet they might be similar in their qualitative behavior, that is,
when one is active the other is also active, when one is large the
other is large, when one is growing the other is growing, and so
on. If one of the signals is inactive and the other is active the
connection is decreased, so we implement a sort of AND operation, that is,  if both are active
the first term in equation 12 becomes 1, hence increasing the weight in
     the first case and decreasing the weight otherwise. At
     the end only those that maintain a recurrent and constant delay
     relation remain. The threshold factors avoid correlations among layers
     that are both inactive. Since when both are inactive the first
     term becomes 0 as both its factors are 0 and the second term is
     also 0. On the other hand, if one is active and the other is
     inactive the "distance" of both signals becomes $\alpha
d^{x,y}_{\tau}(t)$.
Finally if both are active their "similarity" becomes $1-\alpha d^{x,y}_{\tau}(t)$.

The second phase consists of calculating how {\it reliable} the
instantaneous cause-effect relation is within a time interval $[t_0, t_n]$. 
This is calculated by

\begin{equation}
r^{x,y}_{\tau} (t+1) = r^{x,y}_{\tau} (t) + \beta c^{x,y}_{\tau} (t)
\end{equation}
\noindent so that at the end of the training period the measure of the
similarity of both signals will be approximated
by $r^{x,y}_{\tau} (t_N) = \beta\sum_{t=t_0}^{t_n}c^{x,y}_{\tau}(t)$
When a correlation is not "reliable" it will be positive in certain
occasions and negative in others, thus, overall the reliability
measure will be close to 0. From now on ${\bf r}$ denotes the matrix with
components $r^{x,y}_\tau$.
The delays between actions and sensory feedback are reflected on ${\bf
r}$. 
When a cause-effect correlation is reliable (above a
certain threshold) the cause schema becomes the schema that outputs  $o^y(t)$, namely $S^y$.
and the "effect"schema becomes the schema that outputs $o^x(t)$,
namely $S^x$.
Hence, giving rise to the construction of a new predictive schema $P^{X,Y}$.
SBL also constructs its corresponding dual schema $S^{Y,X}$ in parallel to be able to re-construct a previous successful pattern of interaction.
\begin{verbatim}



\end{verbatim}

\begin{center}
----------- FIGURE 12 ABOUT HERE -----------
\end{center}
To facilitate understanding let us first analyze a simplified case.
Consider five schemas, three of which are motor schemas A, B, C and two of which are perceptual schemas 1, 2. 
Suppose there is only a cause-effect relation between motor schema B and perceptual schema 1 with a delay $\tau_2$; 
and a cause-effect relation between motor schema C and perceptual schema 2 with a delay $\tau_1$.  
Figure 12 represents the cause-effect relations among these three motor schemas and these two perceptual schemas. 
To farther exemplify suppose that perceptual schema 1 is sensory feedback from the jaw muscles and 
perceptual schema 2 corresponds to "prey on view field". 
On the other hand, suppose that motor schema A is sidestep,  motor schema B corresponds to ´open mouth´, and motor schema C corresponds to "lunge tongue". 
It should be clear in this example  that there is only a cause-effect relation between the schemas "open mouth" 
and sensory feedback from the jaw with a certain time delay; and the schemas sidestep and "prey on view field" with a probably different time delay.

Returning to our adaptable prey-catching case,
Figure 13 displays ${\bf r}$ under different
conditions (A: after tongue prehension (TP), B: after jaw prehension (JP), and C: after
both tongue and jaw prehension).
Again, the delays between actions and sensory feedback are reflected on ${\bf
r}$. For instance the fourth row in the fourth column of the second block 
in Figure 13A reflects the relationship 
between $jaw\_rec(t)$ and $lm(t)$ as being  delayed by 30 msec (each displayed 
row corresponds to a delay increment of 10 msec, i.e. 0., 10, 20, 30 msec., and so on).
The reader may
have already suspected that some motor programs involve almost
simultaneous activation of several motor units. This raises the
problem where accidental correlations may be
found. One of the ways the system is able to uncouple the "accidental"
pairings occurs when some of the components are active in one motor
program but not in another. A couple of examples may serve to
illustrate the point. The initial accidental pairing $(lu(t), jaw\_rec(t))$
during Jaw Prehension is resolved by the uncoupling during Tongue
Prehension since during TP $lu(t)$ is very small or inexistent whereas
$jaw(t)\_rec$ is very active.
In the same way the initial accidental pairing $(gg(t), jaw\_rec(t))$ is
resolved by the uncoupling during Jaw Prehension, so that although
correlated under one motor program they are de-correlated under another
one. 
Another source of disambiguation comes from the constraint on the
 reliability of the delay between the cause and the effect across
"training samples". As Figure 13C displays, after
the animal has performed both TP and JP a set of predictive schemas
has been formed that represent the internal models of interaction.
They reflect reliable cause-effect relationships that have been found between
schemas output ports. For instance, the following predictive schemas
have been formed:  $P^{DISTX, LU}$ and $P^{JAW\_REC,LM}$.
\begin{verbatim}



\end{verbatim}

\begin{center}
----------- FIGURE 13 ABOUT HERE -----------
\end{center}
\section{The Self-constructive brain is self-growing}
The brain builds itself up, that is, it constructs new patterns of interaction and representation on the run,  as it needs them in order to better understand and build its own interpretation of the environment. 
In order to do so, it grows new specific structures by recruiting dormant overgeneralized structures and instantiating them by specific features that represent specific patterns of interaction with the environment. 
In this regard, prey-catching gives rise to many different extremely relevant (for survival) patterns of interaction with the environment and hence, it is a very good testbed to analyze the construction of representations of patterns of interaction which we here formalize as the construction of new schemas. 

The next two subsections describe firstly schema construction in learning to detour, and secondly the generalized constructive architecture which will be also later applied to learning to snap. 
We hereby describe and formalize schema construction which is a central component of the SCB since it is the mechanism that allows the incremental and dynamic building of representations of relevant patterns of interaction with the environment. Hence, allowing the successful construction of the animat's reality. 

\subsection{Schema construction during Learning to Detour}
As it was already explained in section 4.3.1, 
Corbacho and Arbib (1995) provided a schema-based model of learning to detour in Rana computatrix and we here generalize and formalize this behavior within the SBL constructive architecture. 
In this regard, as already expressed, many predictive schemas imply a sensory or sensorimotor map $S^x$ which is affected by a motor schema $S^y$.
The associations $(S^y, S^x)$ are learned through cause-effect relations already explained in section 5.1 in this paper. 
For instance, the activity pattern in $mhm(t)$ can be affected by different
motor and pre-motor schemas and a different predictive schema is learned for each of these cause-effect relations. 
In this regard, during learning to detour a new predictive schema is constructed as already explained in section 4.3.1.
During the learning process there is a discontinuity, that is, a surprise arises.
When the frog is relatively close to the end of the barrier, a sidestep action suddenly takes the frog out of the barrier and leaves the prey in the open field. 
This immediately gives rise to an unexpected pattern of activity in $mhm(t)$, that is the $S^{MHM}$ schema gets unexpectedly activated. 
During that event, given a previously inactive $S^{MHM}$, the activation of the sidestep schema gives rise to an unexpected pattern of activity in $mhm(t)$, and this triggers the cause-effect dynamics 
which define the effect schema $S^x=S^{MHM}$ and the cause schema $S^y=S^{SIDE}$;
and a new predictive schema is constructed, namely
$P^{MHM, SIDE}$
as well as its corresponding dual schema 
$S^{SIDE. MHM}$, both already introduced in section 4.3.1.
The other ingredient to determine the predictive schema is the definition of the context. To define the context,  first the space(s) have to be defined, secondly the specific patterns that determine the specific context. To define the context the systems gathers statistics by the cause-effect learning dynamics (Corbacho, 1997). 
Next section describes the general schema construction machinery and exemplifies it with the learning to detour case.

\subsection{Schema Construction: generalized architecture}
In general, to construct a new schema, three aspects must be determined: when, what, and how;
which determine the components and the configuration for the new schema(s).
First of all, the brain must realize when there is the need to construct (a) new schema(s). 
In this regard, both the incoherence (prediction error) and the performance error will serve as the triggers under specific conditions, as it will be explained below. 
The next step consists of realizing what can take the system back to coherence and closer to the goal.
This implies determining what components (what) and in which topological configuration (how)  will make up the new schema.
Hence, let us first examine in detail the conditions that lead to the construction of a new schema as well as determine
what other component schemas play a role in the constructive process. 
\begin{verbatim}
\end{verbatim}
{\bf Conditions for Schema Construction:} 
As already explained, SBL attempts to maximize coherence, that is, reduce the prediction error, and, at the same time, maximize performance. 
In this regard, structural learning processes are triggered when specific types of incoherences are detected.
When an incoherence arises, SBL  distinguishes between 
an {\it incorrect expectation} (I), 
versus a completely {\it unexpected} event (U).  
In turn, unexpected can be generated in two cases: (U1) by a new predictive cause-effect relation discovered by the cause-effect dynamics (as described in section 5.1); 
and (U2) when there is an incoherence, that is, the predictive response does not match the actual observed response
$(o^x(t+1) - \hat{o}^{x,y}(t+1)) \neq 0$
and, at the same time, either the cause or the effect schemas are inactive,
i.e., $o^y(t)o^x(t+\tau) = 0$. 

In both of those situations, two sub-cases must be further distinguished, namely,
one occurs when the unexpected pattern of
interaction is accompanied with the system getting closer to a goal (case A),
and the second case, when
it is accompanied with the system
getting farther away from a goal (case B).
In the first case, the system should
attempt to "record" the "configuration" that gave rise to that
interaction. 
Whereas in the second case, the system must make sure to avoid
the current pattern of interaction and re-construct some previous successful pattern (or construct alternative strategies (Corbacho, 1997). 
Specifically, in learning to detour 
the system is on the first case (i.e. U1.A: unexpected pattern of interaction accompanied
with the system getting closer to the goal) as it will be farther elaborated below. 
On the other hand, learning after the lesion corresponds to case (U2.B) since the lesion causes a prediction error due to the inactivation of the effect schema. 
In this case, previous successful pattern re-construction is necessary to achieve the goal. 

Returning to the learning to detour case, as explained in sections 4.3.1 and 5.1, the unexpected pattern of activity in $mhm(t)$ triggers the cause-effect constructive machinery, second the cause-effect machinery determines the association $(S^{SIDE}, S^{MHM})$, third new schemas and schema mappings are constructed to represent the successful pattern of interaction. 
Hence, the cause-effect learning dynamics determines the existing schemas implicated in the construction processes of the new schemas, namely: 
the effect schema is instantiated by  $S^x = S^{MHM}$, and the cause schema is instantiated by $S^y=S^{SIDE}$. 
Thus, the new predictive schema $P^{x,y}$ and its new dual schema $S^{y,x}$ components and their topological configuration are determined as it will be specified below.
Also implicated in the construction process are the goal schema $G^{z,x} = G^{PREY, MHM}$
(related to the active effect schema) and the perceptual schema $S^z=S^{PREY}$ (related to the previous goal schema). That is, the chain of related schemas plays an important role in the constructive process as it is shown in figure 14. 
\begin{verbatim}
\end{verbatim}
{\bf Constructing new predictive schema $P^{x,y}$:} 
As already explained in section 3.1, once the effect $S^x$ and the cause schema $S^y$ have been determined, the predictive schema behavioral specification is defined by a mapping $M^{x,y}_P$ parametrized by ${\bf W}^{x,y}_{P} (t)$, such that
\begin{equation} 	
\hat{o}^{x,y}(t+1) = M^{x,y}_P (o^{x}(t), o^{y}(t), o^{v}(t); {\bf W}^{x,y}_{P} (t))
\end{equation} 
The new predictive schema recruits the dormant structure and parametrizes 
${\bf W}^{x,y}_{P} (t))$ accordingly.
Returning to the learning to detour case, this general equation becomes instantiated by
\begin{equation}
	\hat{mhm}(t+1) = M^{MHM,SIDE}_P (mhm(t),\\
 side(t), sor(t);\\
 {\bf W}^{MHM,SIDE}_P)
\end{equation}
where adaptation of ${\bf W}^{MHM,SIDE}_P$ allows to incrementally reduce $(mhm(t+1) - \hat{mhm}(t+1))$. 

Also its dual schema $S^{y,x}$ gets constructed in parallel.
The dual schema requires as input a goal pattern that must be  provided by a goal schema.
In this regard, the goal schema $G^{z,x}$ learned in
previous successful interactions outputs the "goal" activity
pattern for schema $S^x$ as already explained in section 3.2. 
That is, in the learning to detour case, the specific goal activity pattern corresponds to $mhm^*(t+1)$. 
\begin{verbatim}
\end{verbatim}
{\bf Constructing new dual schema $S^{y,x}$:} 
This new schema's role is to
produce the "optimal" pattern of activity in the cause schema $S^y$ that will in turn give rise, 
though the cause-effect dynamics, to the goal pattern of activity in the effect schema $S^x$  
(that is $mhm^*(t+1)$ in the case of learning to detour).
The components of the new schema $S^{y,x}$ are as follows:
\\
\indent $<Iplist>$: 
The first input port is of the same type and connected to
the output port of the effect schema $S^x$, i.e. $i^{y,x}_{1}(t) = o^x(t)$, 
the second input port is of the same type and connected to the output port of
Goal schema $G^{z,x}$, i.e. $i^{y,x}_{2}(t) = {o}^{*z,x}(t + 1)$.
The third input port corresponds to {\it context} information; i. e. $i^{y,x}_3(t)=o^{v}(t)$
\\
\indent $<Oplist>$: the output port is of the same type as $S^{y}$ input port
and connected to the adapted cause schema $S^{y'}$ (defined below).
\\
\indent $<Behavior>$: the system must produce a modulatory activity pattern input to the adapted cause schema  $S^{y'}$ (defined below) such that it will make the pattern $o^x(t)$ in the effect schema $S^{x}$ get closer to the goal pattern ${o}^{*z,x}(t+1)$ set by the goal schema $G^{z,x}$, that is
\begin{equation}
			o^{y,x}(t) = M^{y,x} (o^x(t), {o}^{*z,x}(t + 1), o^v(t); {\bf W}^{y,x}(t))
\end{equation}
This is analogous to learning the inverse model in distal supervised learning (Jordan \& Rumelhart, 1992). 
That is, as already indicated, in the learning to detour case, the schema $S^{SIDE,MHM}$ tries to diminish the error in the distal sensorimotor space $(mhm^*(t+1) - mhm(t+1)$. 
\begin{equation}
	o^{SIDE,MHM}(t) = M^{SIDE,MHM} (mhm(t), \\
mhm^*(t+1), sor(t); \\
{\bf W}^{SIDE,MHM}(t))
\end{equation}
The cause schema $S^y$ receives a new modulatory input from the dual schema $S^{y,x}$
since the dual schema outputs an optimal pattern of activity for $S^y$  under a specific context.
Hence, the cause schema is adapted to instantiate this new modulatory input under a specific pattern of interaction. These modulatory loops have been shown to have already evolved in lower vertebrates (Ewert, et al., 1999; Ewert, et al., 2001; Ewert, et al., 2006). 
\begin{verbatim}



\end{verbatim}

\begin{center}
----------- FIGURE 14 ABOUT HERE -----------
\end{center}
Thus, in the learning to detour case, the dual schema creates a modulatory input pattern of activity $o^{SIDE,MHM}(t)$, that becomes input to the $S^{SIDESTEP}$ schema. 
In this regard, the $S^{SIDESTEP}$ coordinated control motor program was previously activated by the $S^{BUMP-AVOID}$ schema and now also receives another input activation. 
In this regard, every schema receives different modulatory input channels that get particularly instantiated during the schema construction process.
\begin{verbatim}
\end{verbatim}
{\bf Adapting Schema $S^y$ into $S^{y'}$:}  
Cause schema $S^y$ is adapted to instantiate
a new modulatory input port from the dual schema (represented by dashed lines in Figures 14 and 15). 
In turn, it adapts its
behavioral specification as now it has to take into account this instantiated
modulatory signal. 
\\
\indent $<Iplist>$: They are the same as in $S^y$ (i.e. k input ports) in
addition to a newly instantiated modulatory input port. The modulatory input port is of the same
type and connects to the output port of  the dual schema $S^{y,x}$ 
(hence carrying the modulatory term),  
$i^{y'}_{k+1}(t) = o^{y,x}(t)$
\\
\indent $<Oplist>$: They are the same as in $S^y$. 
\\
\indent $<Behavior>$: The behavioral specification remains the same 
as in  $S^y$  but two extra
assignment expressions are instantiated at the very end of the behavioral
specification to take into account the new modulatory input signal
that serves to reconstruct the pattern of activity proven to be successful in the
past in similar conditions, namely,   

  \[
    \o^{y'}(t) =\left\{
                \begin{array}{ll}
                  i^{y'}_{k+1}(t)  & \mbox{if $i^{y'}_{k+1}(t) \neq 0$}\\
                  o^y(t)  & \mbox{otherwise}\\
                \end{array}
              \right.
  \]
That is, if there is no modulatory input, the output port assignment remains as in the original schema $S^{y}$. Yet, if there is any modulatory input, then the output port assignment corresponds to the input objective modulatory input \footnote{As already expressed, for the sake of clarity and simplification for this article, we assume that only one predictive and its corresponding dual schema are instantiated for each specific active motor schema. Hence,  avoiding issues of integration (linear on nonlinear). Thus, $\hat{o}^x(t) = \hat{o}^{x,y}(t)$, $o^y(t) = o^{y,x}(t)$ and equation 19 can be simplified to reflect this assumption.}.

Returning to the learning to detour case,
the dual schema creates a modulatory input pattern of activity 
$o^{SIDE,MHM}(t)$, 
that gets integrated in the $S^{SIDE}$ schema to generate $side(t)$.
That is, $S^{SIDE,MHM}$ sends a modulatory activity pattern to $S^{SIDE}$, 
which activates the $S^{SIDE}$ schema under a specific context (i.e. in presence of the barrier signaled by activation of $S^{SOR}$).
This, in turn, 
produces an anticipatory pattern in $\hat{mhm}(t+1)$ due 
to the predictive schema (as displayed in figure 6) although, initially, the motor heading map activity was null 
due to the presence of the stationary barrier (i.e. $mhm(t) = 0$). 

In summary, a global picture of this process of schema construction is depicted in
Figure 14. It includes the predictive schema  $P^{x,y}$ and its associated dual schema $S^{y,x}$,
as well as the involved goal schema $G^{z,x}$. 
As already explained, for the particular case of learning to detour, they are instantiated by
$S^y = S^{SIDE}$, $S^x=S^{MHM}$, and $S^z=S^{PREY}$.
Also, a  global picture of the generalized process of schema(s) construction is depicted in
Figure 15. 
In the next section, we will show how learning after a lesion can also be explained with this very same constructive process. 
\begin{verbatim}



\end{verbatim}

\begin{center}
----------- FIGURE 15 ABOUT HERE -----------
\end{center}
\section{The Self-constructive brain is self-repairing}
The brain has an outstanding capacity to recover from many different types of injuries. We present a case in lower vertebrates, specifically how anurans are capable of recovering their snapping capabilities after suffering a lesion of their Hypoglossal nerve as explained in section 4.2.
Corbacho et al. (2005) introduced a partial model of this phenomena within the schema-based modeling framework. We claim that in order to do so,
the brain needs to re-construct a previous  pattern of activity thanks to the construction of internal predictive models. 

In this regard, Innocenti and Nishikawa's (1994) as well as Gleason and Nishikawa (1996) studies on motor learning following hypoglossal transection in toads show that 
the individual learning process is quite idiosyncratic. 
That is, each toad seems to develop its own solution to the 
problem. The adaptation of the motor output may depend highly on the particular features of the prey and the particular physical configuration of the agent itself. 
Thus, the same minimal initial structure will give rise to very different brains if posed on very different environments, since the brains is self-constructive. 
Thus, the system has the tools to build the "vehicle" but which "vehicle" it builds depends on the environment it faces and its previous history of interactions.

The previously presented schema-based learning framework is very helpful as the basis for structural learning, that is,
learning a new topological structure within the brain. 
SBL includes structural learning since it implies the construction of new structures and relations.
This is in contrast to pure parameter learning where the structure is fixed and a set of parameters is adapted. 
Due to space  limitations we will focus on the particular case
when a complex system/agent undergoes lesioning of one of its
subcomponents and the system must learn a structural
change that allows the system to recover its initial functionality.

We, as designers, may know the effect of a particular lesion.
On the other hand, it appears  obvious that natural systems can not have 
any direct knowledge of its own internal relations other than through observation 
of its own dynamics. 
In this regard, we have already discussed that, initially, the system may
not have any relation between pre-motor $S^{LM}$,  $S^{DM}$ and proprioceptive   $S^{JAW\_REC}$  and, only through results of 
interactions, these cause-effect relations can be learned. 
In Corbacho, Nishikawa, Liaw and Arbib (1996) the necessary components and their interrelation already existed 
and learning was just a matter of tuning the existing structures. 
Yet, this poses a combinatorial problem since not all possible projections can be 
contemplated a priori, specially in a hierarchical system, where incremental construction is 
open ended since the environment is dynamic and partially unpredictable. 
In this paper, we 
describe the construction of  new schemas (not determined a priori) under the SBL paradigm.
In this regard, we have already explained in section 5.1 how the cause-effect dynamics trigger 
the construction of internal models by first determining the cause-effect relations among different schema mappings. 
Then, in section 6.2, we have also described how the predictive schema and its dual schemas can be learned to represent these relevant patterns of interaction. 
We claim that these internal models also allow the system to recover after a lesion (Adami, 2006; Bongard, Zykov \& Lipson, 2006; Corbacho, 1997; Corbacho, Nishikawa, Weerasuriya Liaw \& Arbib, 2005b).

\subsection{Pattern (re)-construction to recover from lesion}
We have previously shown how during normal prey capture, the systems builds a predictive schema
$P^{JAW\_REC, LM}$
recording the effects of the pre-motor schema  $S^{LM}$ on 
the proprioceptive schema $S^{JAW\_REC}$. 
That is, the SCB builds internal models of its own dynamics that will allow for better future performance as well as provide for more robust and resilient dynamics (Bongard, Zykov \& Lipson, 2006; Corbacho \& Arbib, 1997a).
Subsequently, after the hypoglossal lesion has taken place, the animal is not able to open the mouth while
attempting to capture the prey.
Hence, the system reaches an incoherence, i. e. a 
prediction error  $(jaw\_rec(t+1) - \hat{jaw\_rec(t+1)}) \neq 0$,
determining  a {\it cause} schema  $S^{LM}$ ($S^y$) that happens to be active
(as expressed by its output port $lm(t) \neq 0$)
and an {\it effect} schema  $S^{JAW\_REC}$ ($S^x$) that remains inactive (i.e. $jaw\_rec(t+1) = 0$).
Thus, corresponding to the case U2.B for schema construction, as explained in section 6.2. 
The system has also performed an unsuccessful interaction as
indicated by the fact that it is getting away from one of its
goals, since the goal schema $G^{PREY,JAW\_REC}$ ($G^{z,y}$) outputs a goal state 
$jaw\_rec^*(t+1)$ reflecting the desired state of $jaw\_rec(t+1)$,
yet, both patterns are not becoming closer.

We suggest that learning after the lesion corresponds to the
re-construction of the activity pattern  similar to that created by
 $S^{LM}$ during successful prey capture before the lesion. 
The dual schema's role would be to reestablish
that activity pattern in  $S^{LM}$ so that the jaw opens and hence 
$S^{JAW\_REC}$ gets properly active.
In this regard, the dual schema $S^{LM , JAW\_REC}$
 (learned in parallel to the predictive $P^{JAW\_REC, LM}$)
outputs the "optimal" activity pattern for the cause
schema $S^{LM}$. 
As already explained for the learning to detour case,
the dual schema creates a modulatory input pattern of activity for the  $S^{LM}$ schema. 
This is so, since every schema receives different modulatory input channels that get 
particularly instantiated during the construction process. 

Figure 16 provides the detailed schema construction process
with respect to the cause schema $S^y$ and the effect schema $S^x$ for generality reasons.
The specific pattern of interaction during learning to snap, after the lesion, is instantiated by
$S^y =  S^{LM}$, $S^x=  S^{JAW\_REC}$ and $S^z =  S^{PREY}$. 
Figure 16, has the exact same topological structure as that of Figure 15 for the case of 
schema construction in learning to detour. 
Hence, again proving the generality of the constructive architecture presented in this paper. 
\begin{verbatim}



\end{verbatim}

\begin{center}
----------- FIGURE 16 ABOUT HERE -----------
\end{center}
\section {Conclusions}
This paper presents adaptive behavior as emerging  from the self-constructive nature of the brain. 
The brain builds itself up by reflecting on its particular interactions with the environment, 
that is, it constructs its own interpretation of reality through the construction of 
representations of relevant predictive patterns of interaction. 
Thus, we claim that the brain is the architect of its own reality. 
In this regard, we formalize the constructive architecture of the brain within the schema-based learning framework. 
We introduce the predictive (cf. forward internal models) and its dual schemas as active processes capturing relevant patterns of interaction; and we suggest that the brain is composed of  myriads of these patterns as well as their dual predictive internal models associated.
These internal models exist all over the brain and a variety of examples can be found in the literature regarding different brain areas/functionalities (visual, sensorimotor, motor, etc.).
Yet, these only represent the "tip of the iceberg', 
since they are all over all the different brain structures 
because they are the result of a central principle of organization.
The existence of all these predictive internal models in the brain can be observed through many anticipatory activity patterns.
We claim that this anticipatory activity patterns are emergent of this type of constructive architecture as reported in this paper showing anticipatory responses in sensorimotor and motor maps. 
We also claim that a constructive architecture must be in place in order to construct and organize all these predictive patterns of interaction.

This paper suggests that the brain is SCB since it is self-experimental,  self-growing,         and self-repairing.
The brain is self-experimental since to ensure survival the self-constructive brain is an active machine capable of performing experiments of its own interactions with the environment 
as well as capable of mentally simulating the results of those interactions in order to be able to later decide the most optimal course of action. 
Anticipation plays an important role in directing intelligent behavior under a fundamental hypothesis that the brain constructs reality as much as it embodies it. 
The brain is also self-growing, since it dynamically and incrementally constructs internal structures in order to incrementally build a model of the world as it gathers statistics from its interactions with the environment.
Finally, the brain is also self-repairing since, to survive, it must also be robust and capable of self-organization and self-repair, that is, the brain is capable of finding ways to repair parts of previously working structures and hence re-construct a previous successful pattern of interaction.

In this paper we have also introduced some of the main components of the SBL constructive architecture and 
claim that they can provide an explanation for 
some of the learning occurring during development as well as after lesioning.
In this regard, toads grow 100 times (e.g. from 0.5 grm to 50 grm) during their lifetime,
so the system has to be intrinsically adaptable and flexible.
Additionally, the recovery after lesioning of the Hypoglossal nerve indicates
 that toads are also 
capable of structural learning by constructing new strategies based on ingredients of past successful interactions.
Hence, we have shown that constructive learning processes underlie the adaptability displayed by anurans and that they can be modeled within the SBL framework.

\subsection {Discussion}
Many researchers have clarified that intelligence resides in the circular relationship between the brain of an individual organism, its body, and the environment (Beer, 1995; Chiel \& Beer, 1997; Nolfi \& Floreano, 2000; Nolfi, Ikegami, \& Tani, 2008; Pfeifer \& Scheier, 1999; Tani, 1996; Varela, Thompson, \& Rosch, 1991). 
Also, an increasing number of research directions emphasize the embodied agent framework (Brooks \& Stein, 1994; Husbands, 2009; Lund, 2014). 
In this regard, different cognitive theories also emphasize the dynamic and interactive role of the brain and the environment  (Barandian, \& Moreno, 2006; Clark, 1997; Clark, \& Grush, 1999; 
Di Paolo, Barandiaran, Beaton \& Buhrmann, 2013;
Nolfi \& Tani, 1999; van Duijn, Keijzer, \& Franken, 2006; Vakarelov, 2011).

This paper definitely emphasizes the importance of the closed interaction loop between the agent and its environment in a dynamical way. 
Future work will farther emphasize the relation to the rest of the organism. 
The SCB is an intrinsic and inherent part of the overall agent as the agent physical configuration changes throughout the lifetime of the agent and hence the SCB must correspondingly adapt its internal models dynamically. 
In this regard, Innocenti and Nishikawa's (1994) as well as Gleason and Nishikawa (1996) studies on motor learning following hypoglossal transection in toads show that 
the individual learning process is quite idiosyncratic. 
That is, each toad, depending on its physical configuration, 
seems to develop its own solution to the problem.
In order to do so, self-modeling is required within the SCB framework.
Nevertheless,  the first stage of learning to open the jaw appears to be common to all toads and, though there are temporal differences during the learning process, some aspects are common  (e.g. first overshooting of mouth opening). 
After the lesion the toad receives no direct external reinforcement from the environment. 
So we have claimed that an expectation-based strategy is one of the few viable ways the 
problem can be solved (Corbacho, Nishikawa, Liaw, \& Arbib, 1996). 

We claim that the SCB principles (self-experimental, self-growing and self-repairing) are evolutionary principles that have evolved in many different species, not just in mammals. That is, these principles can be clearly observed at different levels of abstraction and simple implementations are found even in lower vertebrates. 
So for instance, anuran's brain capacity for recovering a behavioral pattern of interaction after a critical lesion demonstrates their capacity for self-repair. Self-growing and self-experimental, on the other hand, can also be tested by their ability to learn certain problems such as learning to detour through the use of internal models. 
In this regard, this paper provides evidence in anurans and provides its computational counterpart in Rana Computatrix. 

This paper has also emphasized the individual's self-construction of reality, 
yet the construction of reality has important social implications in higher species (Arbib \& Hesse, 1986; Butz, 2008; Piaget, 1954).
The agent lives in a society and the interaction with other agents plays an important role in the agent dynamics. 
Future work involves the extension to include social interactions. 
In this regard, an increasing body of research deals with the formalization of social interactions. 
For instance, Di Paolo \& De Jaegher (2012) introduce the interactive brain hypothesis in order to map the spectrum of possible relations between social interaction and neuronal processes. This hypothesis, among other things, states that interactive experience and skills play enabling roles in both the development and current function of social brain mechanisms, even in cases where social understanding happens in the absence of immediate interaction. 
In this regard, mirror neurons have been related to what to expect from other agents by direct observation, that is, they are related to the expectation of other agents' actions.
Hence, internal forward models have also been related to mirror neuron systems (Gallese \& Goldman, 1998; Orzop, Kawato \& Arbib, 2013; Orzop, Wolpert \& Kawato, 2005; Tani, Ito, \& Sugita, 2004) and predictive coding (Kilner,  Friston \& Frith, 2007). In turn, mirror neurons have been related to the evolution of language (Arbib, 2002, 2005, 2014).
Also, in the field of autonomous robots,  Steels and Spranger (2008) explain how autonomous robots can construct a body image of themselves and how this internal mental model, in turn, can help in motor control, planning and  in recognizing actions performed by other agents. 

\setcounter{secnumdepth}{0} 

\section{Acnowledments:} The author wishes to acknowledge Michael A. Arbib and Christoph von der Malsburg for very enlightening discussions on brain theory and schema-based learning during the author's Ph.D. residence in the Neural, Informational and Behavioral Sciences (NIBS) program at the University of Southern California (Los Angeles).

\section{References}

 Adami, C. (2006). What do robots dream of? Science 314, 1093-1094. 
\newline
\newline
Atkeson, C. G. (1989). Learning arm kinematics and dynamics. 
Annual Review of Neurosci., 12, 157-183. 
\\
\\
Anderson, C. W. \& Nishikawa, K. C. (1993). 
A Prey Type Dependent Hypoglossal Feedback System
in the Frog Rana Pipiens. Brain Behav. Evol.,  42, 189-196.
\\
\\
Anderson, C. W. \& Nishikawa, K. C. (1996).
The roles of visual and proprioceptive information during motor program 
choice in frogs. J. Comp. Physiol. A, 179, 753-762.
\\
\\
Arbib, M. A. (1972). The Metaphorical Brain. New York: Wiley-Interscience.
\\
\\
Arbib, M. A. (1987). Levels of Modeling of Mechanisms of
Visually guided Behavior. Behavioral and Brain Sciences, 10, 407-465. 
\\
\\
Arbib. M. A. (1995). Schema Theory. In: Handbook of Brain Theory and Neural Networks. (ed. Michael A. Arbib). MIT Press. 
\\
\\
Arbib, M. A. (2002). The mirror system, imitation, and the evolution of language. 
In: Dautenhahn, K. \& Nehaniv, C. L. (eds.) Imitation in animals and artifacts. MIT Press, Cambridge, MA: 229-280.
\\
\\
Arbib, M. A. (2005). From monkey-like action recognition to human language: An evolutionary framework for neurolinguistics. Behavioral and Brain Sciences, 28, 105-167. 
\\
\\
Arbib, M. A. (2014). Co-evolution of human consciousness and language (revisited). J Integr. Neurosci., 13(2), 187-200. 
\\
\\
Arbib, M. A. \& Hesse, M. B. (1986). The construction of reality.  Cambridge University Press. 
\\
\\
Arbib, M. A., Lieblich, I. (1977). Motivational learning of spatial
behavior. In: Systems Neuroscience, (ed. J. Metzler) pp. 119-165. New York: Academic Press.
\\
\\
Ashby, W. R. (1960). Design for a Brain. London, UK: Chapman and Hall.  
\\
\\
Bar, M. (2007). The proactive brain: using analogies and associations to generate predictions.
 Trends Cogn. Sci., 11(7), 280-289.
\\
\\
Barandiaran, X. E., \& Moreno, A. (2006). On What Makes Certain Dynamical Systems Cognitive: 
A Minimally Cognitive Organization Program. Adaptive Behavior, 14(2), 171–185.
\\
\\
Beer, R. (1995). A dynamical systems perspective on agent-environment interaction. 
Artificial Intelligence, 72(1), 173-215.
\\
\\
Bell, C. C., Han, V., Sugawara, Y., \& Grant, K. (1997) Synaptic plasticity in a cerebellum-like
 structure depends on temporal order. Nature, 287, 278-281.
\\
\\
Berkes, P., Orban, G., Lengyel, M., \& Fiser, J. (2011). Spontaneous Cortical Activity Reveals
 Hallmarks of an Optimal Internal Model of the Environment. Science, 331, 83-87. 
\\
\\
Bongard, J. Zykov, V. \& Lipson, H. (2006). Resilient machines through contiuous self-modeling.
 Science, 314, 1118-1121. 
\\
\\
Brooks, R., \& Stein, L. A. (1994). Building brains for bodies. Autonomous Robots, 1, 7-25. 
\\
\\
Bubic, A., von Cramon, D. Y., \& Schubotz, R. I. (2010).  Prediction, cognition and the brain.
Front. Hum. Neurosci., 4(25). doi: 10.3389/fnhum.2010.00025.
\\
\\
Butz, M. V. (2002). Anticipatory learning classifier system, Boston, MA: Kluwer Academic Publisher. 
\\
\\
Butz, M. V. (2008). How and why the brain lays the foundations for a conscious self.
Constructivist Foundations, 4, 1-42. 
\\
\\
Butz, M. V. (2010). Curiosity in learning sensorimotor maps. In: J. Haack, H. Wiese, A. Abraham, \& C. Chiarcos (Eds.). KogWis 2010 - 10. Tagung der Gesellschaft für Kognitionswissenschaft (p.92). Potsdam Cognitive Science Series 2. 
\\
\\
Butz, M. V., Herbort, O., \& Pezzulo, G. (2008). Anticipatory, goal-directed behavior. In G. Pezzulo, M. V.  Butz, C. Castelfranchi,  \& R. Falcone (Eds.) The Challenge of Anticipation: A Unifying Framework for the Analysis and Design of Artificial Cognitive Systems, LNAI 5225 (pp. 85-114).  Berlin Heidelberg: Springer
\\
\\
Butz, M. V.  \& Hoffman, J.  (2002). Anticipations control behavior: Animal behavior in an anticipatory learning classifier system. Adaptive Behavior, 10(2), 75-96.
\\
\\
Castelfranchi, C. (2005). Mind as an anticipatory device: For a theory of expectations. In: M. De
Gregorio, V. Di Maio, M. Frucci, C. Musio (eds.), Proceedings of Brain, Vision, and Artificial
Intelligence. Berlin: Springer, pp. 258-276.
\\
\\
Cagliori, D., Tommasino, P., Sperati, V., \& Baldassarre, G. (2014). Modular and hierarchical brain organization to understand assimilation, accomodation and their relation to autism in reaching tasks: a developmental robotics hypothesis. Adaptive Behavior,  22(5), 304-329. 
\\
\\
Chersi, F., Donnarumman, F. \& Pezzulo, G. (2013). Mental imagery in the navigation domain: a computational model of sensory-motor simulation mechanisms. Adaptive Behavior 21(4), 251-262. 
\\
\\
Chiel, H. \& Beer, R. (1997). The brain has a body: Adaptive behavior emerges from interactions of nervous system, body and environment. Trends in Neurosciences, 20, 553-557.
\\
\\
Clark, A. (1997). Being there: Putting brain, body and world together again. Cambridge, MA: MIT Press.
\\
\\
Clark, A. (2013). Whatever next? Predictive brains, situated agents, and the future of cognitive science. Behavioral and Brain Sciences, 36, 181-204.
\\
\\
Clark, A., \& Grush, R. (1999). Towards a Cognitive Robotics. Adaptive Behavior, 7(1), 5-16.
\\
\\
Collett, T. S. Do toads plan routes? A study of the detour behaviour of Bufo viridis. J. Comp. Physilogy, 146, 261-271. 
\\
\\
Corbacho, F. (1997). Schema-based Learning: Towards a Theory of 
Organization for Adaptive Autonomous Agents. Ph.D. Thesis. University of Southern  California, Los Angeles. 
\\
\\
Corbacho, F. (1998). Schema-based Learning. Artificial Intelligence, 101(1-2), 337-339.
\\
\\
Corbacho, F. \& Arbib, M. A. (1995). Learning to Detour. Adaptive Behavior, 3(4), 419-468.
\\
\\
Corbacho, F. \& Arbib, M. A. (1997a). Towards a Coherence Theory of the Brain and Adaptive systems. Proceedings of the First International Conference on Vision, Recognition, and Action. (ed. Stephen Grossberg). Boston, MA.
\\
\\
Corbacho, F. \& Arbib, M. A. (1997b). Learning Internal Models to Detour. 
Society for Neuroscience. Abs. 7, 624.
\\
\\
Corbacho, F. \& Arbib, M. A. (1997c). Schema-based Learning: Biologically Inspired
Principles of Dynamic Organization. 
Lecture Notes in Computer Science 1240. Springer Verlag: Berlin.
\\
\\
Corbacho, F., Nishikawa, K. C., Liaw, J-S., \& Arbib, M. A. (1996). 
An Expectation-based Model of Adaptable and Flexible Prey-Catching in Anurans. 
Society for Neuroscience. Abs. 22: 644.
\\
\\
Corbacho, F. J.; Nishikawa, K. C.; Weerasuriya, A., Liaw, J.S.; Arbib, M. A., (2005a). Schema-based learning of adaptable and flexible prey-catching in anurans. I. The basic architecture. Biological Cybernetics, 93(6), 391-409.
\\
\\
Corbacho, F. J.; Nishikawa, K. C.; Weerasuriya, A., Liaw, J.S.; Arbib, M. A., (2005b).
Schema-based learning of adaptable and flexible prey-catching in anurans. II. Learning after lesioning. Biological Cybernetics, 93(6), 410-425.
\\
\\
Craik, K. (1943). The Nature of Explanation. Cambridge University Press. 
\\
\\
Cruse, H., \& Steinkuehler, U. (1993). Solution of the direct and inverse kinematic problems by a commom algorithm based on the main multiple computations. Biological Cybernetics, 69, 345-351. 
\\
\\
De Ridder, D., Verplaetse, J. \& Vanneste, S. (2013). The predictive brain and the “free will” illusion. Front. Psychol. 4(131). doi: 103389/fpsyg.2013.00131. 
\\
\\
Declerck, G. (2013). Why motor simulation can not explain affordance perception. Adaptive Behavior Journal, 21(4),286-298
\\
\\
Desmurget, M., \& Grafton, S. (2000). Forward modeling allows feedback control for fast reaching movements. Trends in Cognitive Sciences, V. 4(11), 
\\
\\
Di Nuovo, A. G., Marocco, D., Di Nuovo, S., \& Cangelosi, A. (2013). Autonomous learning in humanoid robotics through mental imagery. Neural Networks, 41, 147-155. 
\\
\\
Di Paolo, E. A. \& De Jaegher, H. (2012). The interactive brain hypothesis. Frontiers in Human Neuroscience, 6(163), 1-16. 
\\
\\
Di Paolo, E. A., Barandiaran, X. E., Beaton, M. \& Buhrmann, T. (2013). Learning to perceive in the sensorimotor approach: Piaget´s theory of equilibration interpreted dynamically. Frontiers in Human Neuroscience, 8(551), 1-16. 
\\
\\
Drescher, G. L. (1991). Made-up minds: A constructivist approach to artificial intelligence. MIT Press.
\\
\\
 Driskell, J. E., Copper, C., \& Moran, A. (1994). Does mental practice enhance performance? Journal of Applied Psychology, 79(4), 481-492.
\\
\\
 Droulesz, J., \& Berthoz, A. (1991). A neural network model of sensonitopic maps with predictive short-term properties. Proceedings of the National Academy of Sciences, U.S.A., 88, 9653-9657. 
\\
\\
Downing, K. L. (2009). Predictive models in the brain. Connection Science, 21(1), 39-74. 
\\
\\
Duhamel, J.R., Colby, C. L., \& Goldberg, M. E. (1992) The updating of the representation of visual space in parietal cortex by intended eye movements. Science, 255, 90–92
\\
\\
Ewert, J.-P. (1997) Neural correlates of key stimulus and releasing
 mechanisms: a case study and two concepts. Trends in Neuroscience, 20(8):
332-339.
\\
\\
 Ewert J.-P., Buxbaum-Conradi H., Glagow M., Röttgen A., Sch\"urg-Pfeiffer E., \& Schwippert W.W. (1999). Forebrain and midbrain structures involved in prey-catching behaviour of toads: stimulus-response mediating circuits and their modulating loops. European Journal of Morphology 37, 111-115.
\\
\\
 Ewert, J.-P., Buxbaum-Conradi, H., Dreisvogt, F., Glagow, M.,  Merkel-Harf, C. Rottgen, A., Sch\"urg-Pfeifer, E., \& Schwippert, W. W. (2001). Neural modulation of visuomotor functions underlying prey-catching behaviour in anurans: perception, attention, motor performance, learning. Comparative Biochemestry and Physiology Part A, 128:417-461.
\\
\\
 Ewert J.-P., \& Schwippert W.W. (2006) Modulation of visual perception and action by forebrain structures and their interactions in amphibians. In: Levin E.D. (ed.) Neurotransmitter Interactions and Cognitive Function. Birkhäuser, Basel, pp.99-136.
\\
\\
Flanagan, J., \& Johansson, R. (2003). Action plans used in action observation. Nature, 424, 769-771. 
\\
\\
 Flanagan, J. R., Vetter, P., Johansson, R. S. \& Wolpert, D. M. (2003). Prediction precedes control in motor learning. Curr. Biol. 13, 146-150. 
\\
\\
 Flanagan, J. R., \& Wing, A. M. (1997). The role of internal models in motion planning and contriol: evidence from grip force adjustments during movements of hand-held loads. J. Neurosci. 17, 1519-1528. 
\\
\\
 Gallese, V., \& Goldman, A. (1998). Mirror neurons and the simulation theory of the mind-reading. Trends Cognitive Sci. 2, 493-501.
\\
\\
 Garcia, C. E., Prett, D. M., \& Morari, M. (1989). Model Predictive
Control: Theory and Practice- a survey. Automatica, 25, 335-348.
\\
\\
 Gentili, R., Han, C. E., Schweighofer, N., \& Papaxanthis, C. (2010). Motor learning without doing: Trial-by-trial improvement in motor performance during mental training. J. Neurophysiol., 104, 774-783. 
\\
\\
 Gleason, T., \& Nishikawa, K. C. (1996). The effect of practice regime on motor
learning following hypoglossal transection in toads (Bufo marinus). American Soologist, 36, 116A.
\\
\\
 Gilbert, D.  T. \& Wilson, T. D. (2007). Prospection: experiencing the future. Science, 317, 1351-1354.
\\
\\
 Gregory, R. L. (1967). Will seeing machines have illusions? In Machine Intelligence 1 (N. L. Collins and D. Michie eds.). Oliver \& Boyd. 
\\
\\
Grobstein, P. (1992). Directed movement in the frog; motor choice, spatial
representation, free will?. In Neurobiology of motor programme selection 
(Eds. J. Kien, C. R. McCrohan, \& W. Winlow). Pergamon Press: Oxford.
\\
\\
 Grossberg, S. (2009). Cortical and subcortical predictive dynamics and learning during perception, cognition, emotion and action. Phil. Trans. R. Soc. B., 364, 1223-1234. 
\\
\\
 Guazzelli, A, Corbacho, F., Bota, M. \& Arbib, M. A. (1998). An implementation
of the Taxon-Affordance System for Spatial Navigation.  Adaptive Behavior, 
6(4), 435-471.  
\\
\\
 Haruno, M., Wolpert, D. M. \& Kawato, M. (2001). MOSAIC model for sensorimotor learning and control. Neural Computation, 13, 2201-2220. 
\\
\\
 Hassabis, D. \& Maguire, E. A. (2009). The construction system of the brain.  Phil. Trans. R. Soc. B. 364, 1263-1271.
\\
\\
 Hawkins, J. (2004). On Intelligence. Henry Holt and Company, New York, 2004.
\\
\\
 Hesslow, G. (2002). Conscious thought as simulation of behavior and perception. Trends in Cognitive Sciences, 6, 242-247.
\\
\\
Hesslow, G. (2012). The current status of the simulation theory of cognition. Brain Research, 1428, 71-79. 
\\
\\
 Hinton, G. E., Dayan, P., Frey, B. J. \& Neal, R. (1995). 
The wake-sleep algorithm for unsupervised Neural Networks. Science, 268, 1158-1161. 
\\
\\
Hoffmann, J., Berner, M., Butz, M. V., Herbort, O., Kiesel, A., Kunde, W. \& Lenhard, A. (2007). Explorations of anticipatory behavioral control (ABC): a report from the cognitive psychology unit of the University of Wurzburg. Cogn. Process, 8, 133-142. 
\\
\\
 Husbands, P. (2009).  Never mind the Iguana, What about the Tortoise? Models in Adaptive Behavior. Adaptive Behavior, 17(4), 320-324. 
\\
\\
 Innocenti, C. M. \& Nishikawa, K. C. (1994). Motor Learning in Toads (Bufo marinus) following  Bilateral Hypoglossal Nerve Transection. American Zoologist, 34, 56A.
\\
\\
Jeannerod, M. (2001). Neural simulation of action: a unifying mechanism for motor cognition. Neuroimage, 14, 103-109. 
\\
\\
Johnson-Laird, X, Y.  (1983). Mental models: Towards a cognitive science of language, inference, and consciousness. Cambridge. Cambridege University Press and Harvard University Press. 
\\
\\
 Jordan, M.I. (1983). Mental practice. Unpublished dissertation proposal, Center for Human Information Processing, University of California, San Diego.
\\
\\
 Jordan, M. I. \& Rumelhart, D. (1992). Forward models:
Supervised Learning with a Distal Teacher. Cognitive Science, 16, 307-354.
\\
\\
Kalman, R. E. (1960). A new approach to linear filtering and
prediction theory. Trans. ASME J. Basic Eng., 82, 35-45.
\\
\\
Kawato, M. (1999) Internal models for motor control and trajectory planning. Curr. Opin. Neurobiol. 9, 718-727
\\
\\
Kawato, M. (1990a). Feedback-error-learning neural network for supervised
learning. In R. Eckmiller (Ed.), Advanced neural computers (pp. 365-372). Amsterdam: North-Holland.
\\
\\
Kawato, M. (1990b). Computational schemes and neural network models for formation and control of multijoint arm trajectory. In W.T. Miller, III, R.S. Sutton, \& P.J. Werbos
(Eds.), Neural networks for control. Cambridge, MA: MIT Press.
\\
\\
Kawato, M., Furukawa, K., \& Suzuki, R. (1987). A hierarchical neural network model for control and learning of voluntary movement. Biological Cybernetics, 57, 169-185.
\\
\\
 Kilner, J. M., Friston, K. J., \& Frith, C. D. (2007). Predictive coding: an account of the mirror neuron system. Cognitive Processing 8, 159-166.
\\
\\
 Kossyln, S. M., Gani, G. \& Thompson, W. L. (2001). Neural foundations of imagery. Nature Reviews Neurosci. 2(9), 635-642.
\\
\\
 Lallee, S. \&  Dominey, P. F. (2013). Multi-modal convergence maps: from body schema and self-representation to mental imagery. Adaptive Behavior 21(4), 274-285
\\
\\
 Lieblich, I. \& Arbib, M. A. (1982). Multiple representations of space underlying behavior. Brain Behav. Sci., 5: 627-659.
\\
\\
 Lund, H. H. (2014). Building bodies and brains. Adaptive Behavior, 22(6), 392-395. 
\\
\\
 Lyons D. \& Arbib, M. A. (1989). A formal model of
Distributed computation for Scjhema-based robot control. IEEE
J. Robotics and Automation, 5:280-293.
\\
\\
 Mehta, B. \& Schaal, S. (2002).  Forward models in visuomotor control. Journal of Neurophysiology 88, 942-953.
\\
\\
Merfeld, D. Zupan, L. \& Peterka, R. J. (1999). Humans use
Internal Models to Estimate gravity and linear acceleration. Nature,
398:615.
\\
\\
Metcalfe, J.S., Chen, L.C., Kopp, M.A., Jeka, J.J., \& Clark, J.E. (2001). Beyond postural sway reduction: Do newly walking infants couple to a driving somatosensory stimulus? The First World Congress: Motor Development and Learning in Infancy, Amsterdam, The Netherlands. 
\\
\\
 Metcalfe, J.S. \& Clark, J.E. (2000). Coordinating perception and action: the role of sensory information in the exploration of posture. Biannual conference of the International Society for Infant Studies, Brighton, UK.
\\
\\
Miall, R.C., Weir, D. J., Wolpert, D. M. \& Stein, J. F. (1993) Is the cerebellum a Smith predictor? J. Mot. Behav. 25, 203–216
\\
\\
Miall, R. C. (2003). Connecting mirror neurons and forward models. Neuroreport 14, 2135-2137.
\\
\\
Miall, R.C. \& Wolpert, D. M. (1996). Forward Models for Physiological Motor Control. Neural Networks. 9(8): 1265-1279. 
\\
\\
 Miceli, M. \& Castelfranchi, C. (2002). The Mind and the Future. The (Negative) Power of Expectations. Theory \& Psychology, 12(3): 335-366. 
\\
\\
Mohan, V., Sandini, G. \& Morasso, P. (2014). A Neural Framework for Organization and Flexible Utilization of Episodic Memory in Cumulatively Learning Baby Humanoids. Neural Computation 26, 2692-2734. 
\\
\\
 Munoz, D. P., Pellison, D., \& Guitton, D. (1991). Movement of neural activity on the superior colliculus motor map during gaze shifts. Science, 251, 1358-1360. 
\\
\\
 Mussa-Ivaldi F. A., Bizzi E. (2000). Motor learning through the combination of primitives. Philos Trans R Soc Lond B Biol Sci 355(1404):1755-69. 
\\
\\
 Mussa-Ivaldi, F. A. (1999). Modular features of motor control and learning. Current Opinion in Neurobiology, vol. 9, pp. 713-717.
\\
\\
 Nishikawa, K. C., and Gans, C. (1992). The role of hypoglossal sensory feedback during feeding in 
the Marine toad, Bufo marinus. J. Experimental Zoology, 264:245-252.
\\
\\
Nishikawa, K. C., Anderson, C. W., Deban, S. M., \& O'Reilly, J. C. (1992). The Evolution of Neural 
Circuits Controlling Feeding Behavior in Frogs. Brain Behav. Evol. 40:125-140.
\\
\\
Nolfi, S., \& Floreano, D. (2000). Evolutionary robotics: The biology, intelligence, and technology of self-organizing machines. Cambridge, MA: MIT Press. 
\\
\\
 Nolfi, S., Ikegami, T. \& Tani, J. (2008). Behavior and Mind as a Compex Adaptive system. Adaptive Behavior, 16(2), 101-103. 
\\
\\
 Nolfi, S., \& Tani, J. (1999). Extracting regularities in space and time through a cascade of prediction networks: The case of a mobile robot navigating in a structured environment. Connection Science, 11, 125-148. 
\\
\\
 Oztop, E., Wolpert, D. \& Kawato, M., (2005). Mental state inference using visual control parameters. Cognitive Brain Research 22, 129-151. 
\\
\\
 Oztop, E., Kawato, M., \& Arbib, M. A. (2013). Mirror neurons: Functions, mechanisms and models. Neurosci. Letters 540, 43-55.
\\
\\
 Pezzulo, G. (2008). Coordinating with the future: The Anticipatory Nature of Representation. Minds and Machines, 18, 179-225. 
\\
\\
 Pezzulo, G., Butz, M. V., Sigaud, O., \& Baldassarra, G.  (2009). Anticipatory behavior in adaptive learning systems. LNCS 5499.
\\
\\
Pezzulo, G., Candidi, M., Dindo, H., \& Barca, L. (2013). Action simulation in the human brain: Twelve questions. New Ideas in Psychology, 1-21.
\\
\\
 Pfeifer, R. \& Scheier, C. (1999). Understanding intelligence. Cambridge, MA: MIT Press.
\\
\\
 Piaget, J. (1954). The construction of reality in the child. Ballantine. New York. 
\\
\\
 Rao, R. P. N. \& Ballard, D. H. (1999). 
Predictive coding in the visual cortex: a functional interpretation of some extra-classical 
receptive-field effects. Nature Neuroscience, 2(1):79-87.
\\
\\
 Raos, V., Evangeliou, M.N., \& Savaki, H. E. (2007). Mental simulation of action in the service of action perception. The Journal of Neuroscience 27, 12675-12683. 
\\
\\
 Redon, C. et al. (1991) Proprioceptive control of goal directed movements in man studied by means of vibratory muscle tendon stimulation. J. Mot. Behav. 23, 101–108. 
\\
\\
Rosen, R. (1985). Anticipatory systems. Oxford: Pergamon Press. 
\\
\\
 Sanchez-Montanes, M., \& Corbacho, F. (2004). A new Information Processing Measure for Adaptive Complex Systems. 
IEEE Trans. Neural Netw., 15(4), 917-927. 
\\
\\
Schutz, B. \& Prinz, W. (2007). Prospective coding in event representation. Cognitive Processes, 8, 93-102. 
\\
\\
 Shadmehr, R., \& Mussa-Ivaldi, F. (1994). Adaptive representation of dynamics during learning of a motor task. Journal of Neuroscience, 14, 3208-3224. 
\\
\\
 Shadmehr, R., Smith, M. A., \& Krakauer, J. W. (2010). Error correction, sensory prediction, and adaptation in motor control. Annual Review of Neuroscience, 33, 89-108. 
\\
\\
Sigaud, O. Butz, M. V. Pezzulo, G. \& Herbort, O. (2013). The anticipatory construction of reality as a central concern for psychology and robotics. New Ideas in Psychology, 31(3), 217-220. 
\\
\\
 Steels, L. \& Spranger, M. (2008). The robot in the mirror. Connection Science, 20(4), 337-358. 
\\
\\
 Suri, R. E., \& Schultz, W. (2001).Temporal Difference Model
 Reproduces Anticipatory Neural Activity. Neural Computation, 13(4), 841-862.
\\
\\
 Sutton, R.S. (1988). Learning to predict by the methods of temporal differences. Machine Learning, 3, 9-44.
\\
\\
 Sutton, R.S. (1990). Integrated architectures for learning, planning, and reacting based on approximating dynamic programming. Proceedings of the Seventh International Conference
on Machine Learning pp. 216-224.
\\
\\
 Sutton, R. S., \& Barto, A. G. (1981). An adaptive network
that constructs and uses an internal model of its world. Cognition and
Brain Theory, 4(3): 217-246.
\\
\\
 Sutton, R. S., \& Barto, A. G. (1998). Reinforcement learning: An introduction. MIT Press. 
\\
\\
 Svensson, H., Thill, S., \& Ziemke, T. (2013). Dreaming of electric sheep? Exploring the functions of dream-like mechanisms in the development of mental imagery simulations. 
Adaptive Behavior 21(4), 222-238.
\\
\\
 Szpunar, K. K., Watson, J. M., \& McDermott, K. B. (2007). Neural substrates of envisioning the future. Proc. Natl. Acad. Sci. U.S.A., 104, 642-647. 
\\
\\
 Tani, J. (1996). Model-based Learning for mobile robot navigation from the dynamical systems perspective. IEEE Transactions on Systems, Man, and Cybernetics B, 26(3), 421-436.
\\
\\
 Tani, J., Ito, M., \& Sugita, Y. (2004). Self-organization of distributedly represented multiple behavior schemata in a mirror system: reviews of robot experiments using RNNPB, Neural Networks, 17, 1273-1289. 
\\
\\
Vakarelov, O. (2011). The cognitive agent: Overcoming informational limits. Adaptive Behavior, 19(2), 83-100. 
\\
\\
 van Duijn, M., Keijzer, F. \& Franken, D. (2006). Principles of Minimal Cognition: Casting Cognition as Sensorimotor Coordination. Adaptive Behavior, 14(2), 157-170.
\\
\\
 Varela, F., Thompson, E., \& Rosch, E. (1991). The embodied mind. Cambridge, MA: MIT Press. 
\\
\\
 von Holst, E. \& Mittelstaedt, H. (1950) Das reafferenzprinzip. Wechelwirkung Zwischen Zentralnerven system und peripherie. Naturwis. 37, 464–476.
\\
\\
 Weerasuriya, A., \& Mills, J. (1996). Long Term Consequences of Hypoglossal Nerve Lesions on Anuran Prey Capture. Workshop on Sensorimotor Coordination. Sedona, AR.
\\
\\
 Wada, Y., \& Kawato, M. (1993). A neural network model for arm trajectory formation using forward and inverse dynamics models. Neural Networks, 6, 919-932.
\\
\\
 Wolpert, D. M., \& Flanagan, J. (2001). Motor prediction. Current Biology, 11, 729-732. 
\\
\\
 Wolpert, D. M., Ghahramani, Z., \& Jordan, M. I. (1995). An
Internal model for sensorimotor integration. Science, 269:1880-1882.
\\
\\
Wolpert, D.M., Miall, R. C., \& Kawato, M.  (1998) Internal models in the cerebellum. Trends Cognit. Sci., 2, 338–347
\\
\\
Wolpert, D. M., \& Kawato, M. (1998). Multiple paired forward and inverse models for motor control. Neural Networks 11, 1317-1329.

\newpage

\section{Figures and Table Captions}
~\\
~\\
{\bf Figure 1.} Behavioral observations showing the frog interacting with different pailing fence barrier configurations in the presence of prey. 
(A) Approach to prey with 10-cm. wide barrier interposed.
(B) Approach to prey with 20-cm. wide barrier interposed: first trial (numbers indicate the succession of the movements).
(C) Approach to prey with 20-cm. wide barrier interposed after four learning trials.
Arrowheads indicate position and orientation of the frog following a single continuous movement, after which the frog pauses. 
~\\
~\\
{\bf Figure 2.}  Electromyographic activity of the mandibular depressors (DM) and levators (LM) during snapping behavior before(Left) and after (Right) bilateral transection of the ramus hypoglossus (HG). 
The onset of activity is nearly simultaneous both before and after surgery. Before transection, the depressors reach their peak of activity on average 86.7 ms earlier than the levators . After transection, the peak of activity of the levators occurs nearly simultaneously with the peak of activity of the depressors. [From Nishikawa and Roth 1991; reprinted with permission.]
~\\
~\\
{\bf Figure 3.} Initial seed schema-based architecture for learning to detour. For the purpose of this paper, pay special attention to the prey recognition and stationary object recognition (SOR) schemas which get integrated in the motor heading map (MHM). Also the bumping avoidance schema is implemented by tactile input sending a reorienting  signal to the MHM. 
~\\
~\\
{\bf Figure 4.} Activity patterns in the motor heading map $mhm(t)$,  displaying the motor heading map integration. (A) for the 10-cm wide barrier. (B) for the 20-cm wide barrier.
~\\
~\\
{\bf Figure 5.} Information flow for learning to detour. This diagram provides a schematic representation of the schema activation flow and how information patterns flow in the schema network. The adaptation modules to reduce the prediction error $(mhm(t+1) - \hat{mhm}(t+1))$ and the performance error $(mhm^*(t+1) - mhm(t+1))$ are displayed aside the signal flow to avoid cluttering the image for clarity reasons. Delay lines are represented by D. 
~\\
~\\
{\bf Figure 6.} Anticipatory activity pattern in $\hat{mhm}(t+1)$ as a result of the activation of the predictive schema $P^{MHM,SIDE}$ in the presence of prey and in the context of the barrier, i.e. the schema $S^{SOR}$ is activated and setting up the context. 
~\\
~\\
{\bf Figure 7.} Initial seed schema-based architecture for learning to snap. Namely, control circuitry for depressor mandibularis, levator mandibularis, tongue protractor, tongue retractor, head ventro-flexion and lunging. Some of the neural network structure is simplified to allow for inclusion of lunge and head down flexion MPGs as well as the corresponding sensory feedback. For the purpose of this paper pay special attention to premotor DM and LM in charge of mouth opening. 
~\\
~\\
{\bf Figure 8} Temporal pattern of activity for each schema instance.
Motor pattern when feeding on prey (A) before and (B) right after the bilateral Hypoglossal transection.   The left column reflects the prey catching pattern for TP and the right column reflects the JP pattern.
The first rows display the temporal
pattern of activity corresponding to the different perceptual schemas:
PREY\_REC, PREY\_NORM, HG\_REC, JAW\_REC, TONGUE\_REC, DISTX, DISTY,
respectively. The next row displays the activity of the interneuron
INT1. The following rows display the temporal pattern of activity
corresponding to the different premotor schemas: LU, HE, DM, LM,
GG, HO respectively.
Notice no activity in the third row (HG\_REC) after the HG lesion (B) which causes the DM and LM activity to be simultaneous, hence, causing the mouth not to open and the tongue not to be active (last two rows). 
~\\
~\\
{\bf Figure 9.} Information flow for learning to snap.
This diagram provides a schematic representation of the schema activation flow and how information patterns flow in the schema network. Notice the same topological configuration as that of figure 5. The adaptation to reduce the prediction error $(jaw\_rec(t+1) - \hat{jaw\_rec}(t+1))$ and the performance error $(jaw\_rec^*(t+1) - jaw\_rec(t+1))$ are displayed outside the signal flow to avoid cluttering the image for clarity reasons. 
~\\
~\\
{\bf Figure 10.}  The first row displays the goal pattern of 
activity $jaw\_rec^*(t+1)$, i.e. the subgoal state 
produced by the goal schema $G^{PREY,JAW\_REC}$ 
(due to the $prey(t)$ activity pattern).
The second row displays the modulatory activity pattern in $o^{LM, JAW\_REC}(t)$ produced by
the activation of the dual schema $S^{LM, JAW\_REC}$. 
In summary these activity patterns are produced by 
 the goal and dual schemas  chain:
 $prey(t) \rightarrow jaw\_rec^*(t+1) \rightarrow o^{LM, JAW\_REC}(t)$.
\\
\\
{\bf Figure 11.}  A. Reconstruction of the "optimal" $lm'(t)$ activity pattern 
due to the projection of the correcting modulatory pattern of activity from the 
constructed dual schema $S^{LM, JAW\_REC}$.
The first row displays the original pattern of activity $lm(t)$, 
the second row displays the optimal modulatory activity pattern $o^{LM, JAW\_REC}(t)$,
and the third row displays, for clarity reasons, the difference between the original pattern and the one produced by the dual schema, namely,  $o^{LM, JAW\_REC}(t) - lm(t)$. 
\\
\\
{\bf Figure 12.} Representation of the cause-effect relations in the simplified example consisting of five schemas, 
three of which are motor schemas A, B, C and two of which are perceptual schemas 1, 2. 
The black squares represent the existence of a cause-effect relation between the respective motor and perceptual schemas with that specific delay $\tau_i$.
On the other hand, the white squares represent the lack of any cause-effect relation between the respective motor and perceptual schemas.
\\
\\
{\bf Figure 13.} Cause-effect parameters {\bf r} after
 learning during the different prey-catching motor programs.
 A. After Tongue Prehension (TP). 
 B. After Jaw Prehension (JP). 
 C. After both TP and JP.
Different 
columns indicate the plausible causes for the corresponding  effects  (sensory feedbacks) by indicating 
the weight (by the size of the square -black positive and white negative). 
Columns are grouped by 
type of sensory feedback in five groups of six. The five different groups are separated by a complete 
line of white blocks. For 
instance the first six columns represent the relation between HG\_REC 
and LU, HE, DM, LM, GG, HO 
respectively. The second six columns represent the relation between
JAW\_REC and LU, HE, DM, LM, 
GG, HO respectively. The other groups of columns correspond to TONGUE\_REC, DISTX and DISTY 
respectively. 
Rows, on the other hand, correspond to different delays starting with 0 msec and with
consecutive increments of 10 msec.
Notice that HG-REC did not "correlate" well with any of the "potential" 
causes. 
This actually reflects the fact that HG-REC is independent (spontaneous activity linked with 
respiratory processes) and has no relation with any other schema (relations represented as full white squares). 
Notice the initial accidental pairing  (LU, JAW-REC) during the Jaw 
Prehension  (B) is resolved by the uncoupling during Tongue Prehension since during TP  LU is very 
small or inexistant whereas JAW-REC is very active. 
In the same way the initial accidental pairing  
(GG, JAW-REC) during TP (A) is resolved by the uncoupling during Jaw Prehension. 
So that after both TP and JP (C) the remaining cause-
effect relations are: (DM, JAW\_REC), (LM, JAW-REC), (GG, TONGUE\_REC), (HO, TONGUE\_REC), 
(LU, DISTX), (LU, DISTY), and (HE, DISTY).
\\
\\
{\bf Figure 14.} Overall process of schema(s) construction for learning to detour.
Construction of the new  predictive schema $P^{MHM, SIDE}$ and its dual schema $S^{SIDE, MHM}$, as well as  transformation of the schema  $S^{SIDE}$ to receive a new modulatory input. 
(A) Already existing schemas that determine the cause and effect schemas.
(B) Goal schemas previously learned.
(C) New predictive and dual schemas and its new topological configuration with respect to the 
pre-existing schemas. 
Dashed lines indicate the new connections formed during the constructive
process. $S^{MHM}$ input to the dual schema $S^{SIDE, MHM}$ is replicated aside not to clutter the image with lines. 
\\
\\
{\bf Figure 15.} Generalized process of schema(s) construction. 
Construction of the new  predictive schema  $P^{x,y}$ and its dual schema $S^{y,x}$.
In turn, cause schema $S^y$ is adapted to instantiate
a modulatory input port from the dual schema (represented by dashed lines).
This in turn adapts its
behavioral specification as now it has to take into account this instantiated
modulatory signal.
\\
\\
{\bf Figure 16.} Overall process of schema(s) construction for learning to snap. 
Notice that this figure has the same topological configuration as that of Figure 15 but instantiated by $S^y = S^{LM}$ and $S^x = S^{JAW\_REC}$.

\newpage
\section{Figures}

\begin{figure}[ht!]
\centering
\includegraphics[width=100mm]{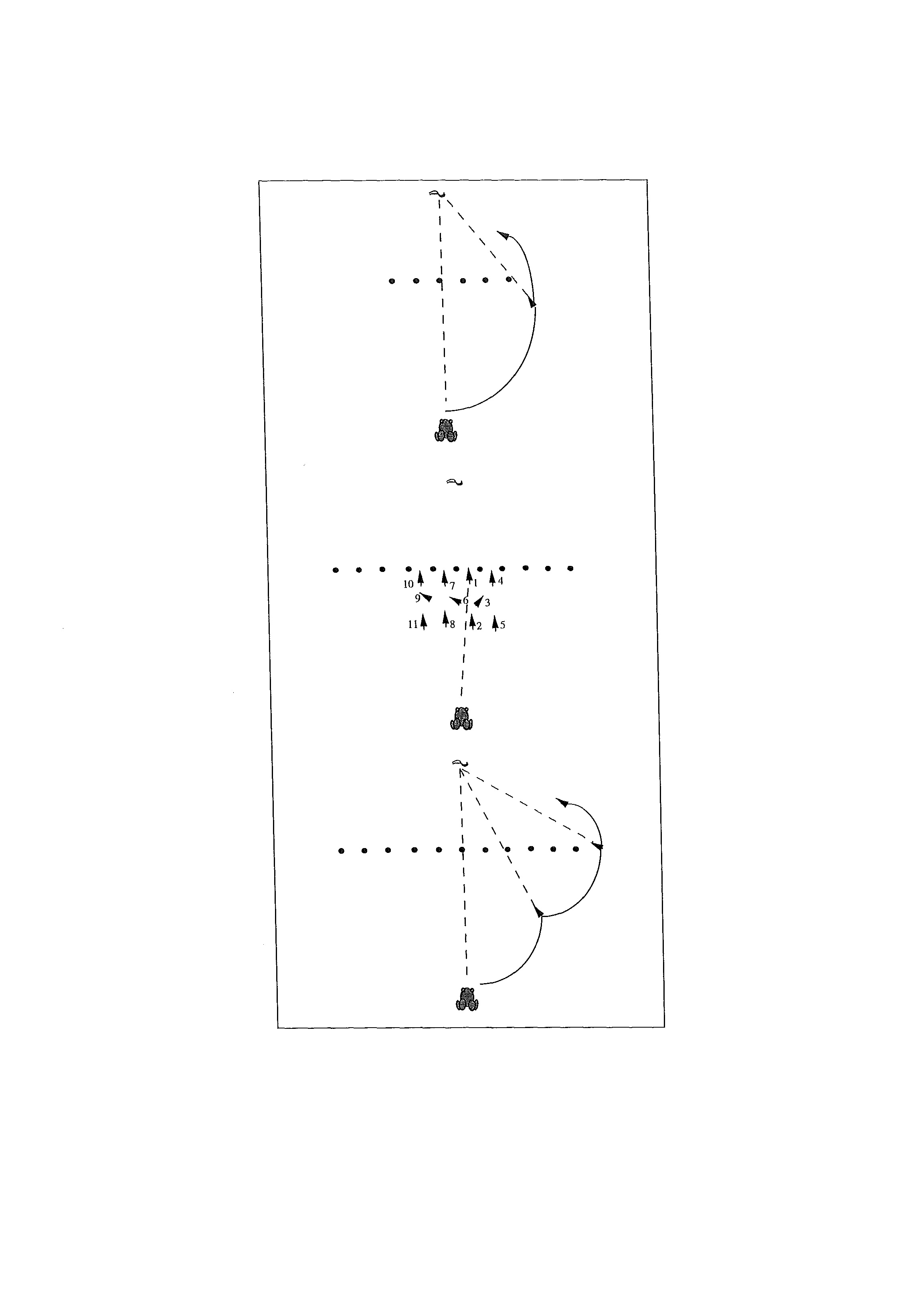}
  \caption{ }
  \label{fig:boat1}
\end{figure}

\begin{figure}[ht!]
\centering
\includegraphics[width=90mm]{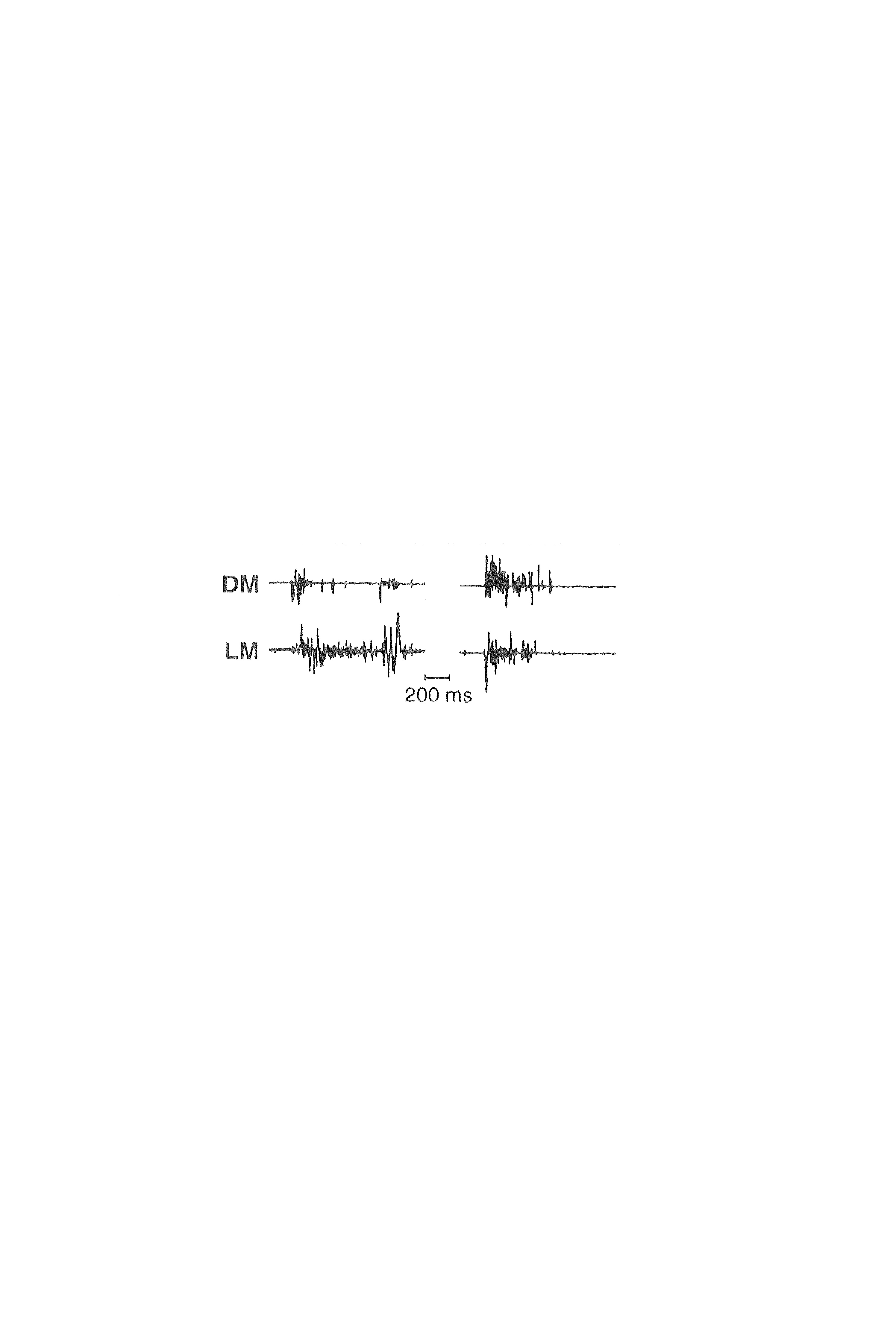}
\caption{    \label{overflow}}
\end{figure}

\begin{figure}[ht!]
\centering
\includegraphics[width=120mm]{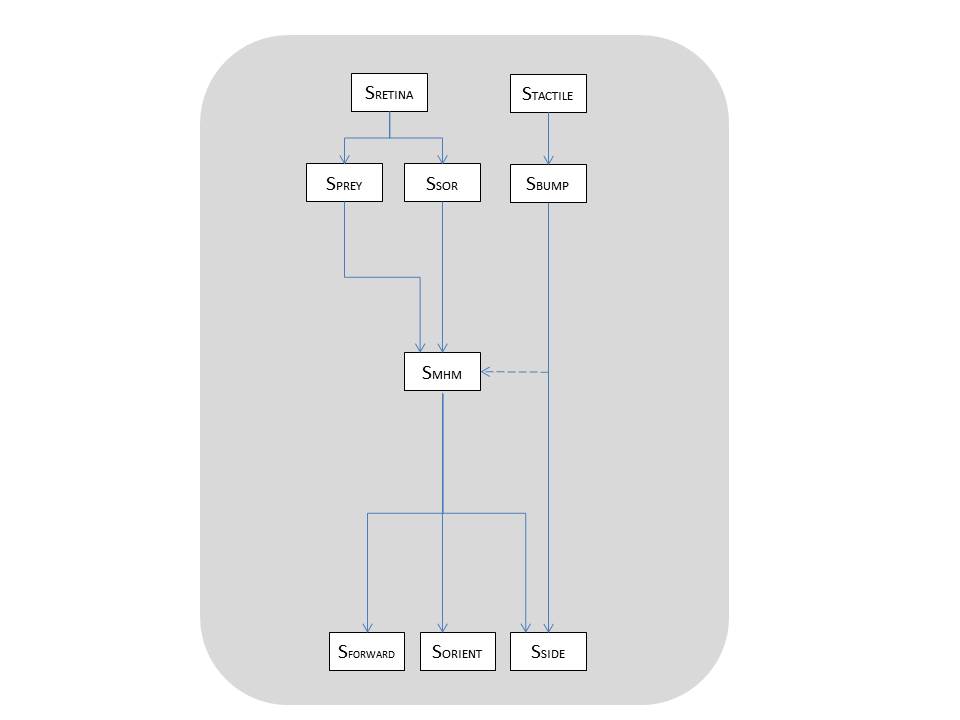}
  \caption{ }
  \label{fig3}
\end{figure}

\newpage

\begin{figure}[ht!]     
  \centering
  \subfloat[A.]{\includegraphics[width=0.5\textwidth]{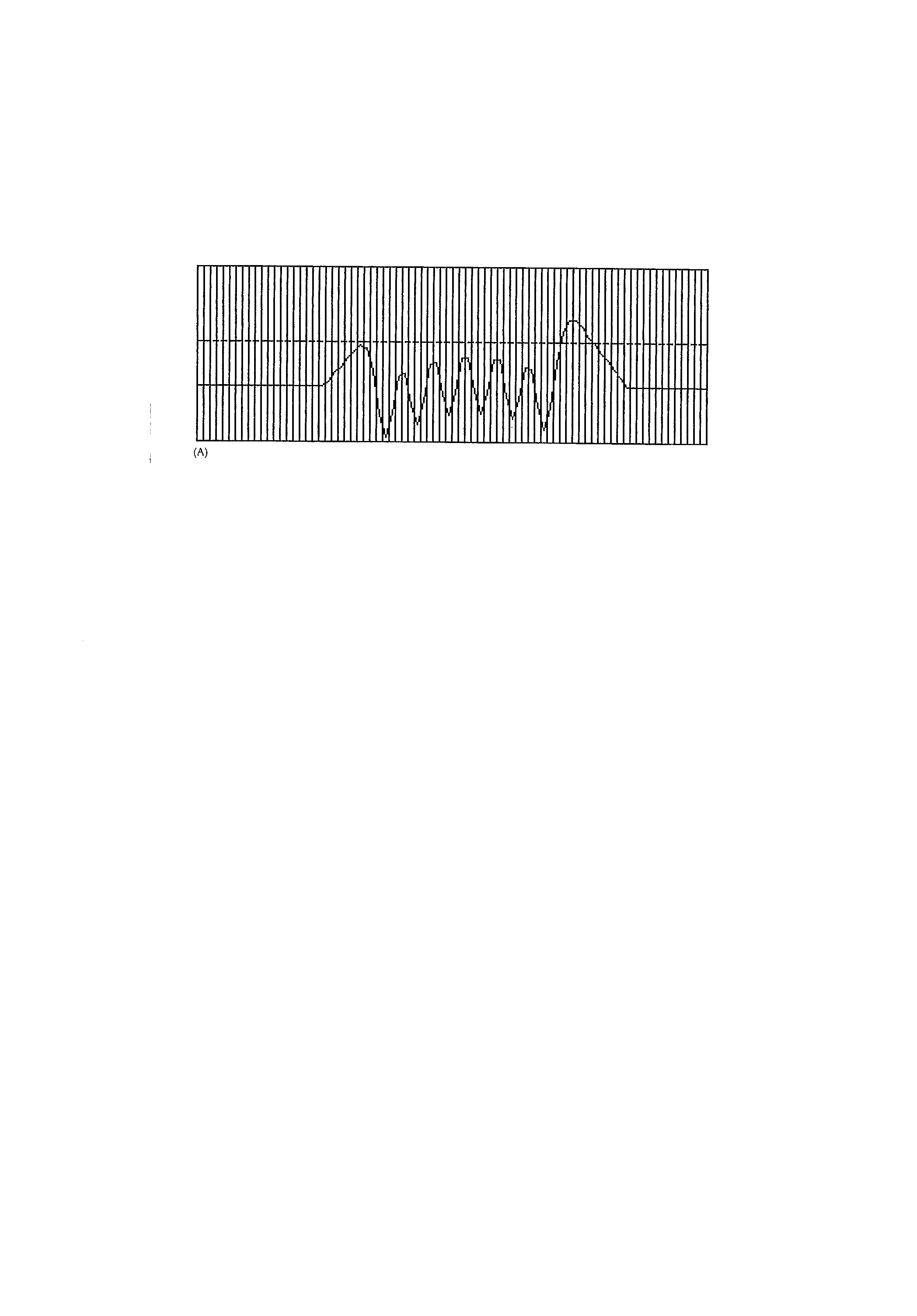}\label{fig:f1}}
  \hfill
  \subfloat[B.]{\includegraphics[width=0.5\textwidth]{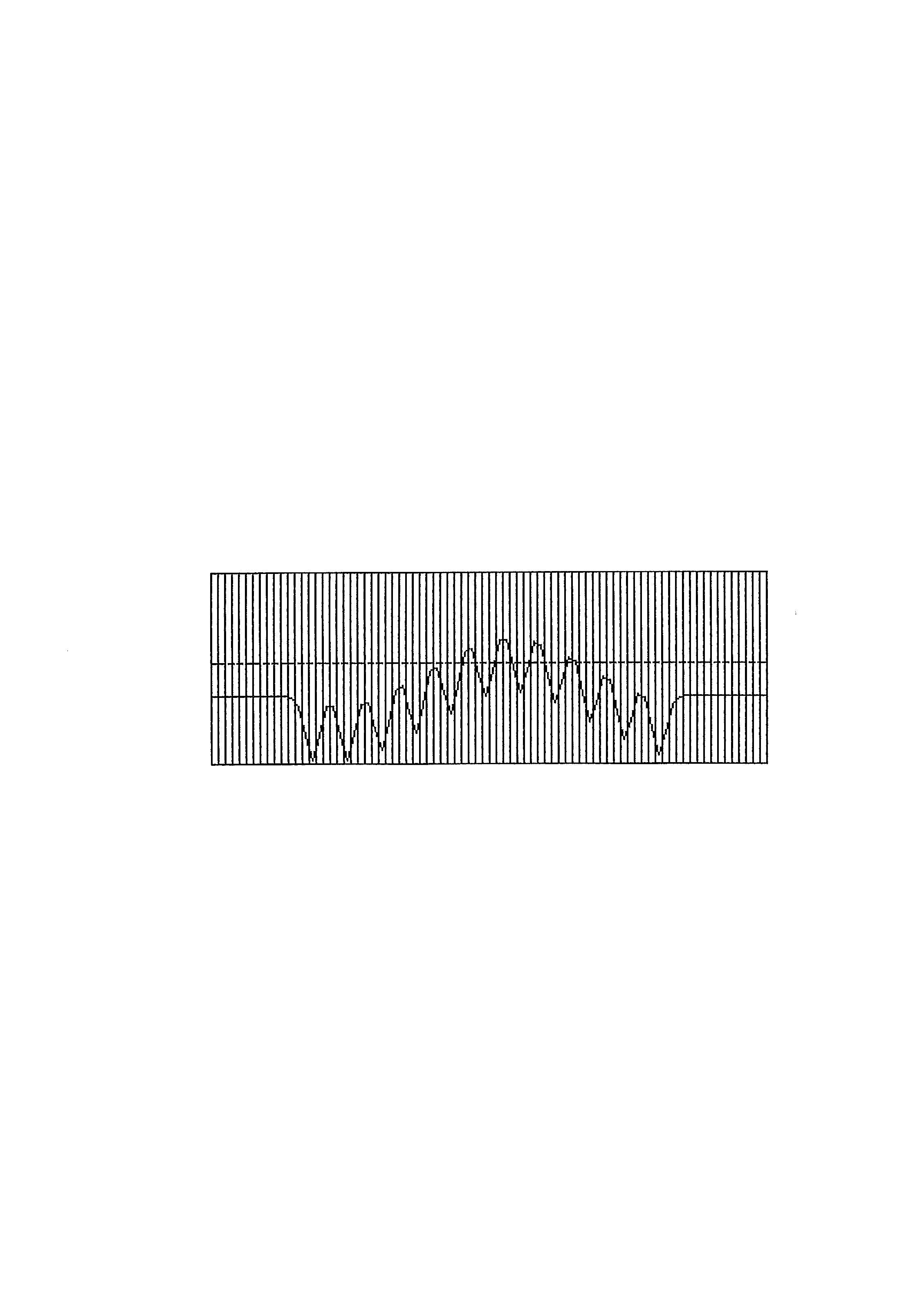}\label{fig:f2}}
  \caption{}
\end{figure}

\begin{figure}[ht!]
\centering
\includegraphics[width=120mm]{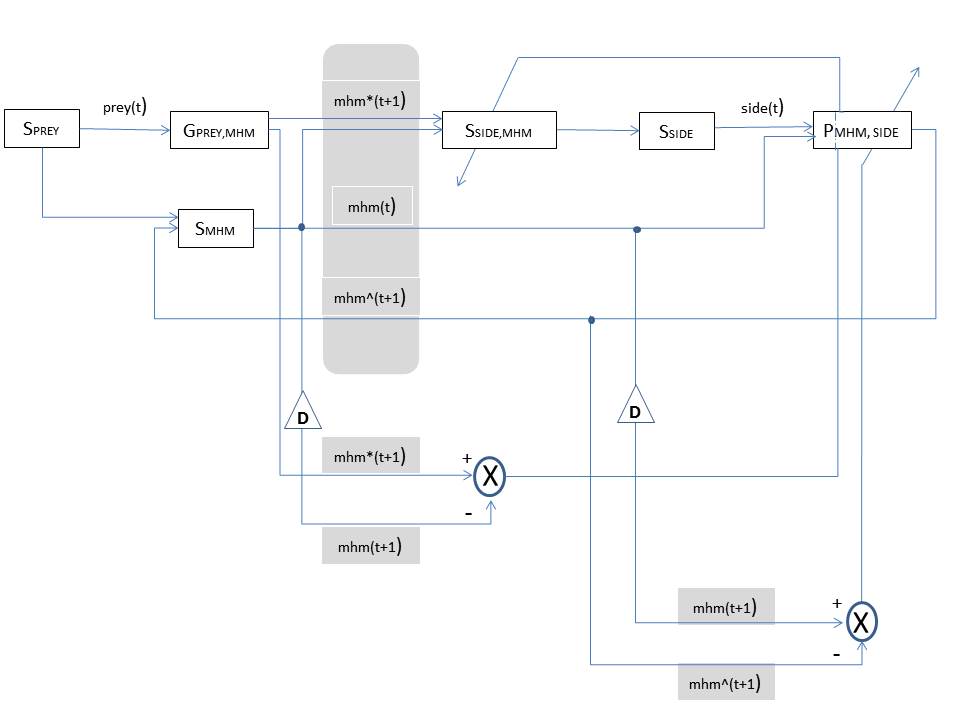}
  \caption{ }
  \label{fig5}
\end{figure}

\begin{figure}[ht!]
\centering
\includegraphics[width=120mm]{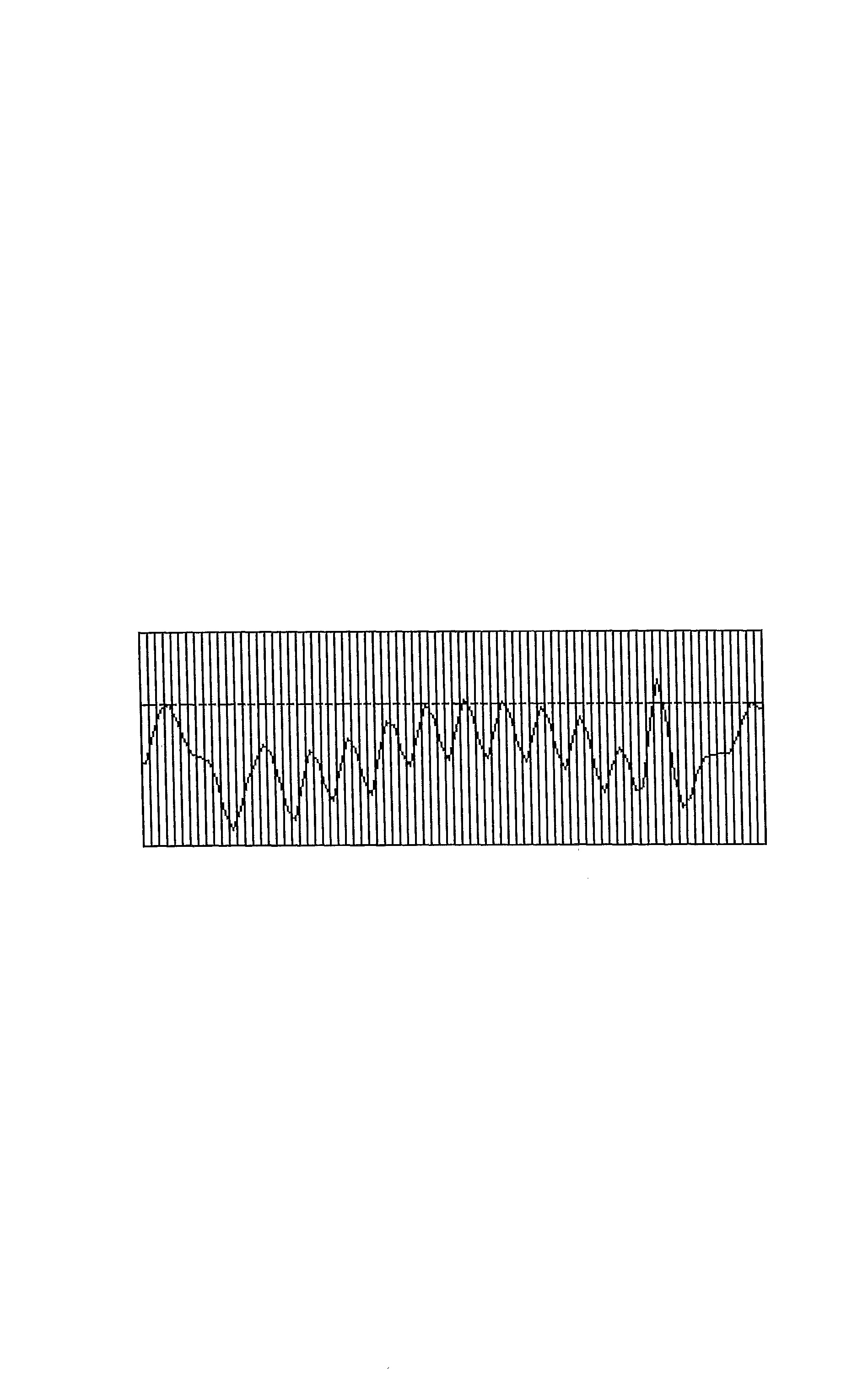}
  \caption{ }
  \label{fig6}
\end{figure}

\begin{figure}[ht!]
\centering
\includegraphics[width=120mm]{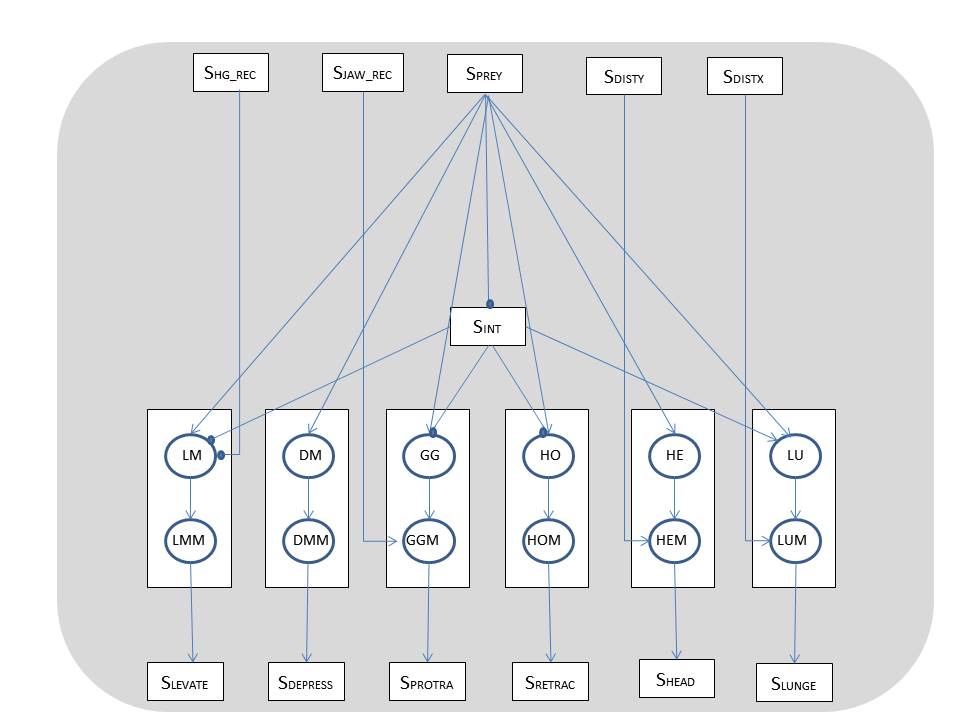}
  \caption{ }
  \label{fig7}
\end{figure}

\newpage
\newpage

\begin{figure}[ht!]     
  \centering
  \subfloat[A.]{\includegraphics[width=0.5\textwidth]{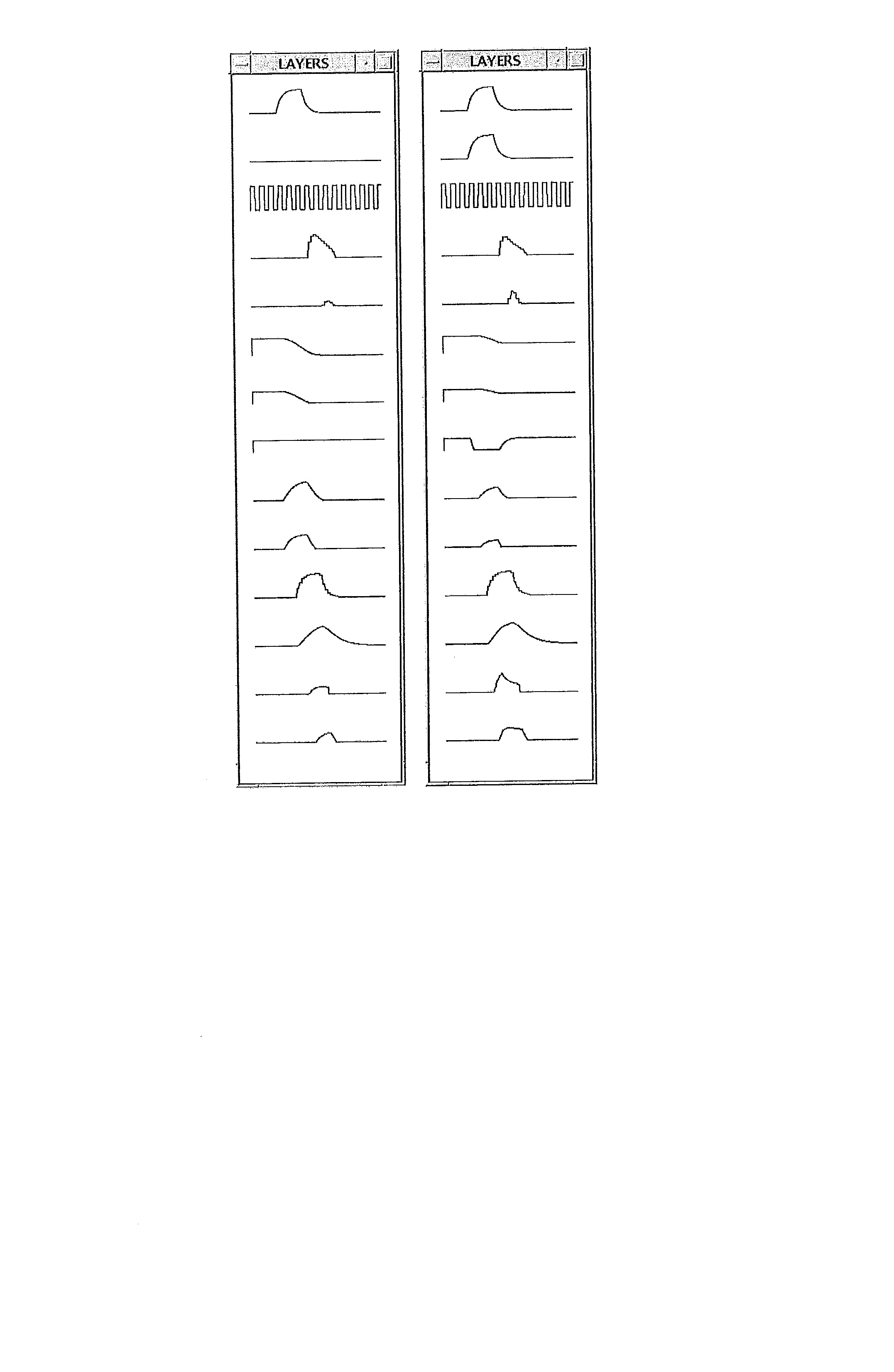}\label{fig:f1}}
  \hfill
  \subfloat[B.]{\includegraphics[width=0.5\textwidth]{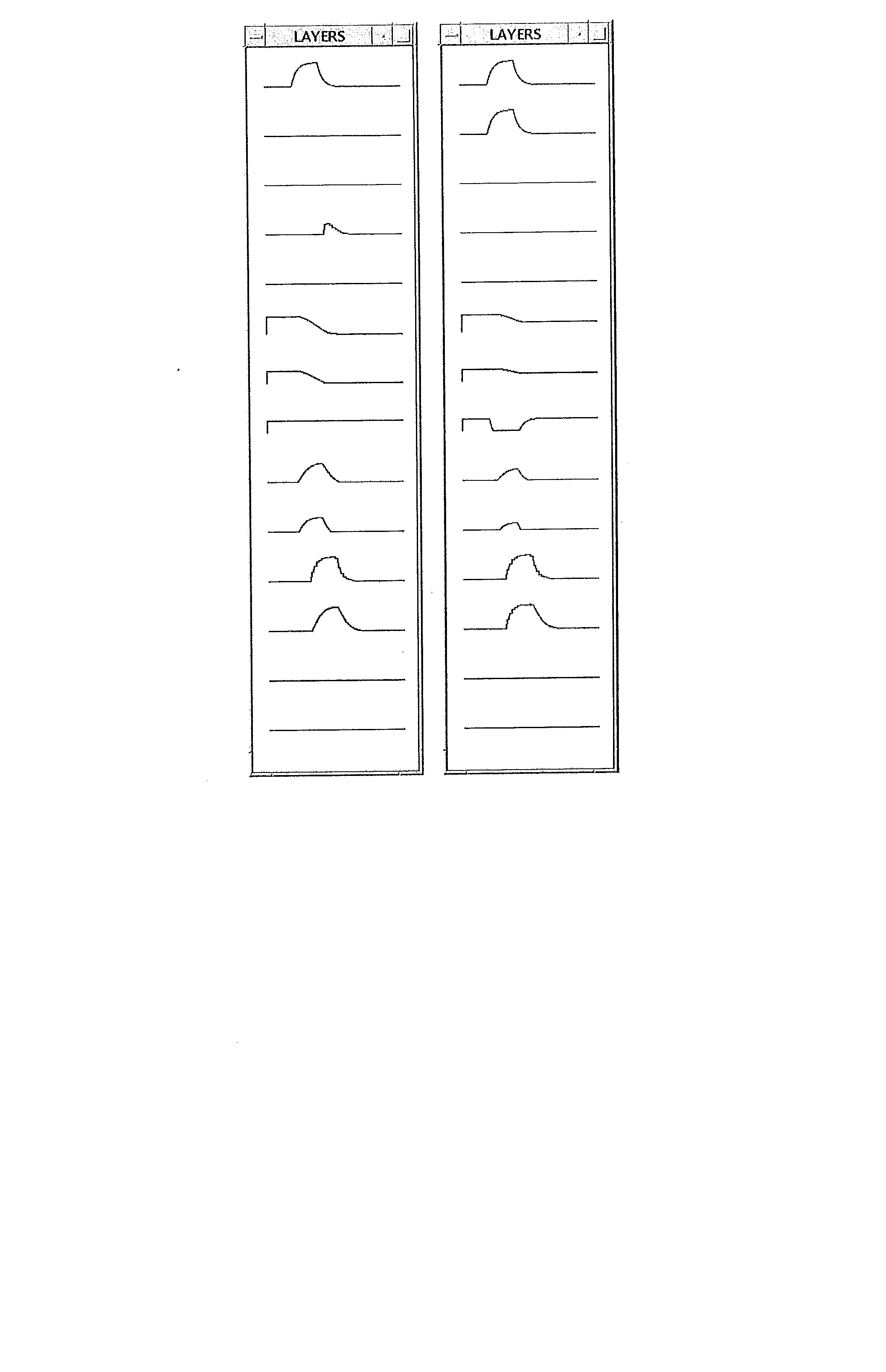}\label{fig:f2}}
  \caption{}
\end{figure}

\begin{figure}[ht!]
\centering
\includegraphics[width=120mm]{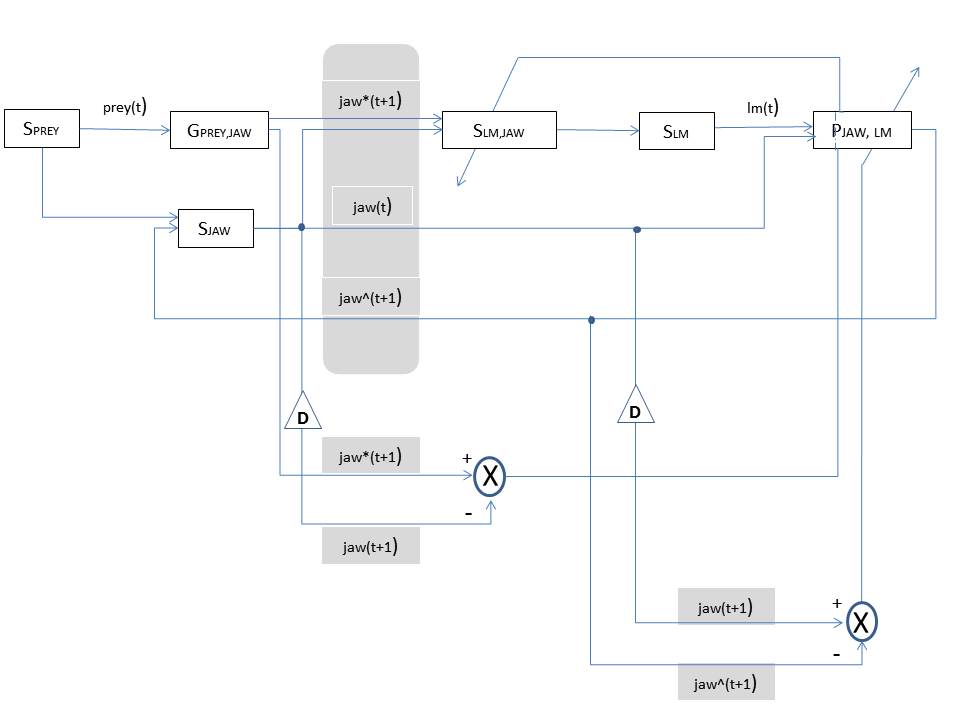}
  \caption{ }
  \label{fig9}
\end{figure}

\begin{figure}[ht!]
\centering
\includegraphics[width=120mm]{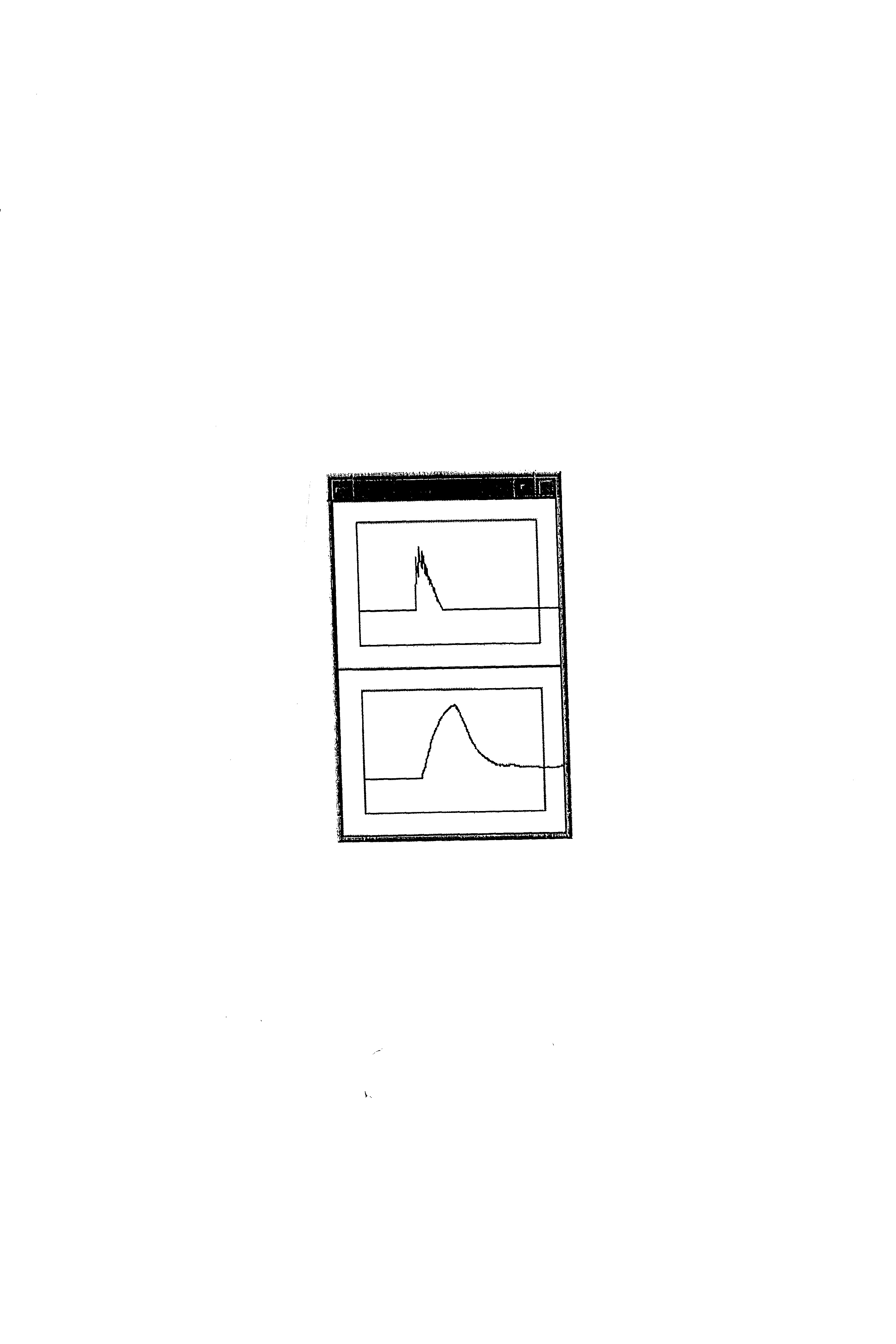}
  \caption{ }
  \label{fig10}
\end{figure}

\begin{figure}[ht!]
\centering
\includegraphics[width=120mm]{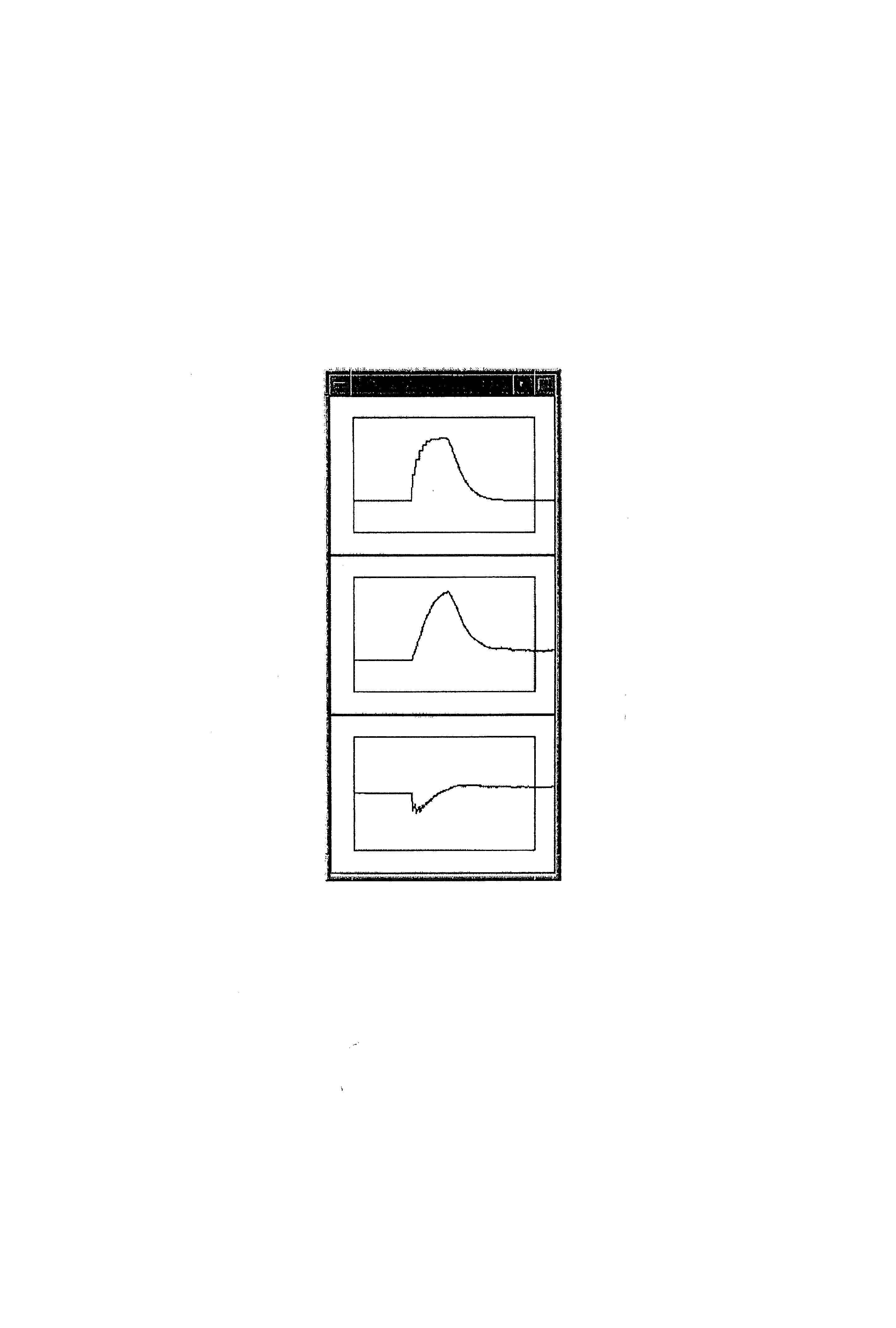}
  \caption{ }
  \label{fig11}
\end{figure}

\begin{figure}[ht!]
\centering
\includegraphics[width=120mm]{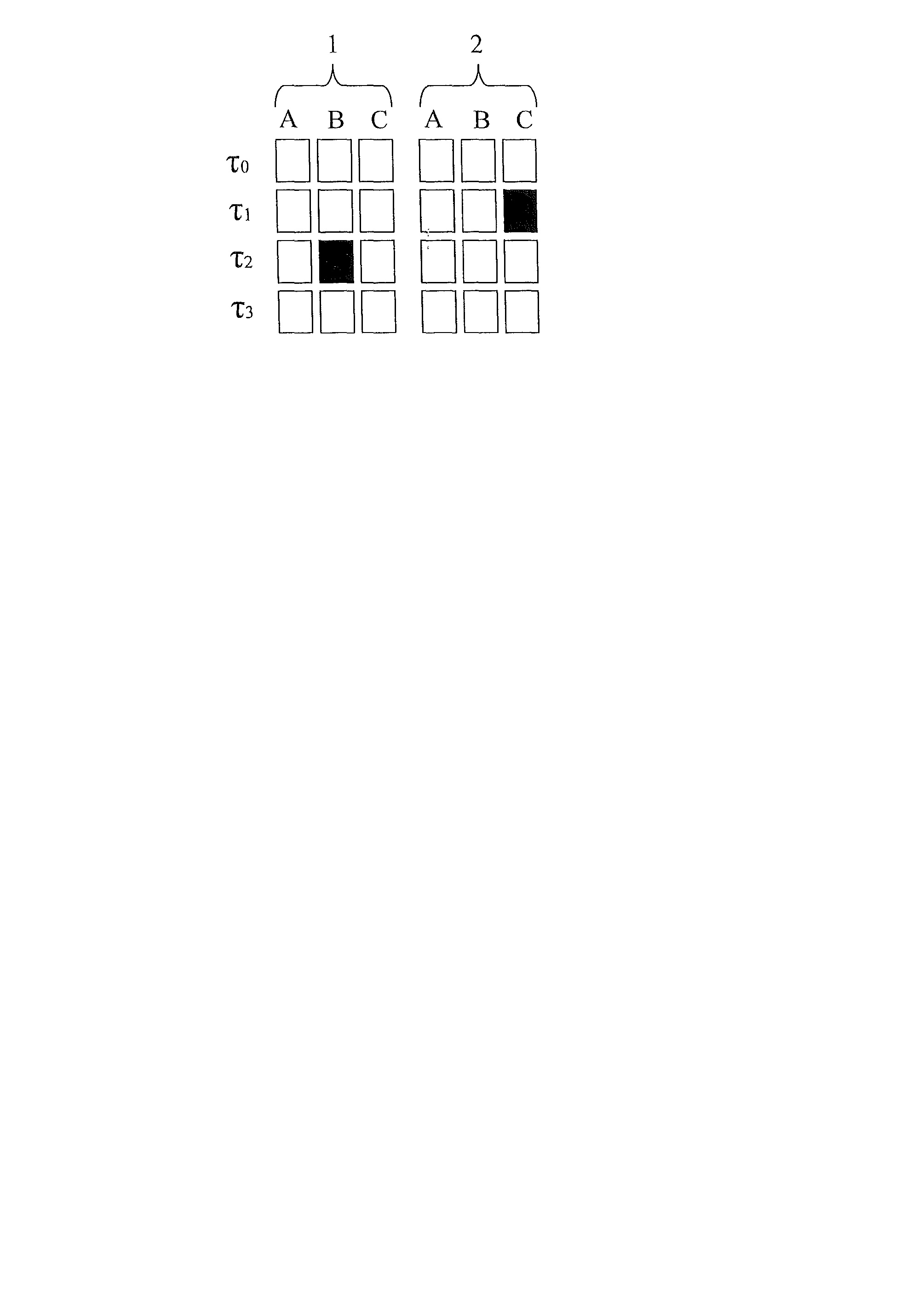}
  \caption{ }
  \label{fig12}
\end{figure}

\begin{figure}[ht!]
\centering
\includegraphics[width=120mm]{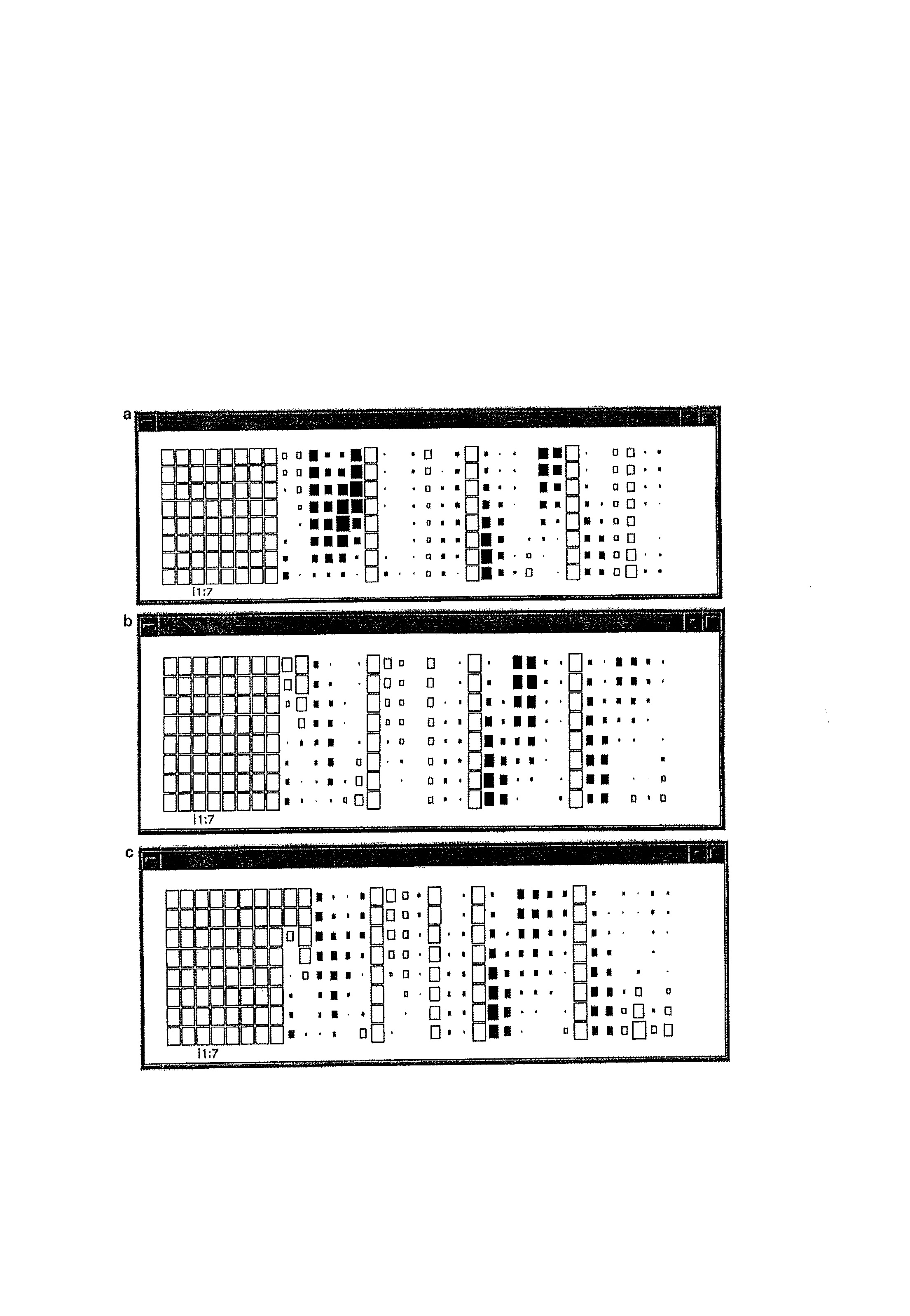}
  \caption{ }
  \label{fig13}
\end{figure}

\begin{figure}[ht!]
\centering
\includegraphics[width=120mm]{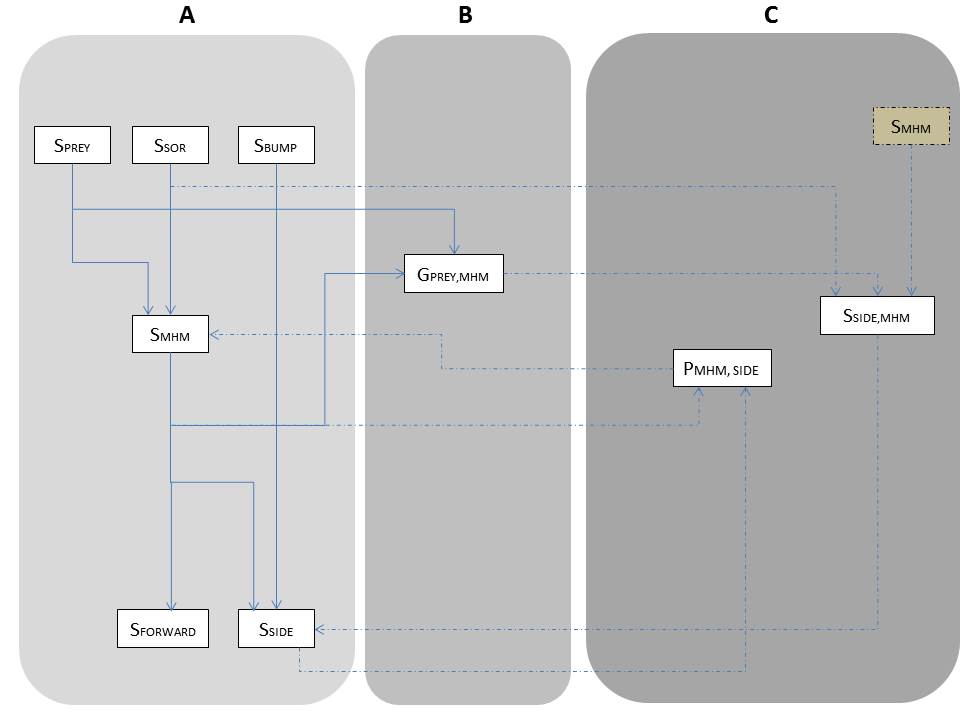}
  \caption{ }
  \label{fig14}
\end{figure}

\begin{figure}[ht!]
\centering
\includegraphics[width=120mm]{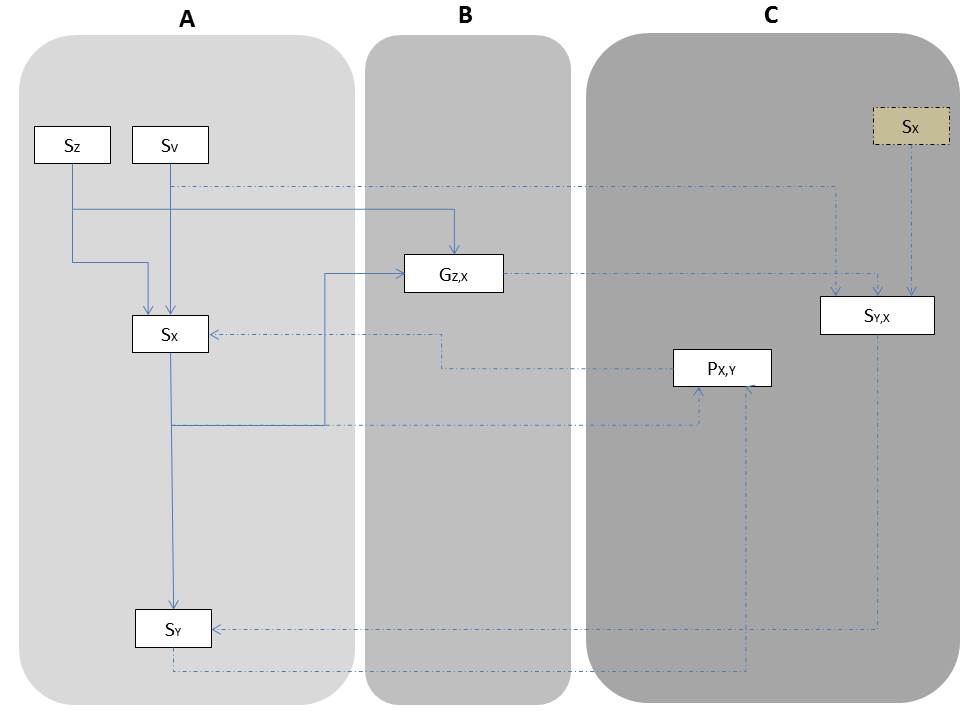}
  \caption{ }
  \label{fig15}
\end{figure}

\begin{figure}[ht!]
\centering
\includegraphics[width=120mm]{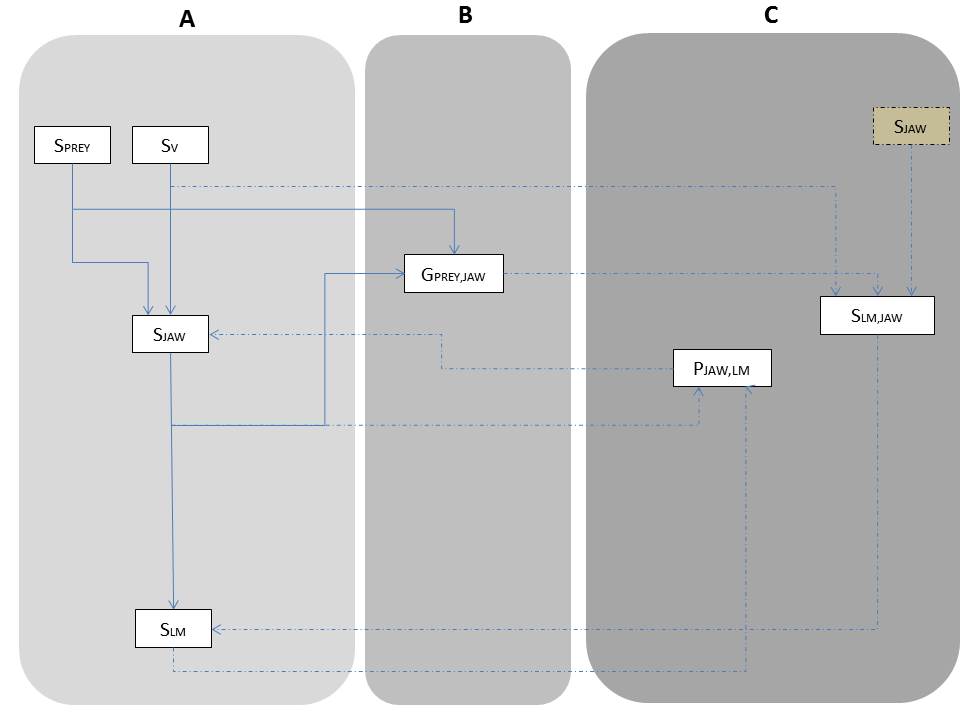}
  \caption{ }
  \label{fig16}
\end{figure}


\end{singlespacing}

\end{document}